\documentclass{article}

\usepackage{times}

\usepackage[scaled=.90]{helvet}
\usepackage{courier}

\usepackage{graphicx} % more modern
\usepackage{subfigure}
\usepackage{amsmath}
\usepackage{xspace}
\usepackage{xcolor}
\usepackage{enumitem}
\usepackage{mathtools}
\usepackage{wrapfig}
\usepackage{caption}
\usepackage{natbib}

\usepackage{algorithm}
\usepackage[noend]{algorithmic}
\usepackage{hyperref}

\usepackage[accepted]{icml2018}

\usepackage{amssymb, amsthm}
\usepackage{tabulary}
\usepackage{booktabs}

\def\equationautorefname~#1\null{Equation~#1\null}
\def\sectionautorefname~#1\null{Section~#1\null}
\def\subsectionautorefname~#1\null{Section~#1\null}
\def\subsubsectionautorefname~#1\null{Section~#1\null}
\makeatletter \newcommand{\ALC@uniqueautorefname}{line} \makeatother

\usepackage[]{color-edits}
\addauthor{md}{defaultcolor}
\addauthor{nj}{olive}
\addauthor{hle}{red}
\addauthor{aa}{blue}

\usepackage[normalem]{ulem}
\renewcommand{\sout}[1]{}

\AtBeginDocument{\colorlet{defaultcolor}{.}}
\newcommand\edit[1]{\textcolor{defaultcolor}{#1}}

\usepackage{comment}

\theoremstyle{plain}
\newtheorem{theorem}{Theorem}

\newcommand{\Ex}{\mathbb{E}}
\newcommand{\indi}[1]{\mathbb{I}\left[#1\right]}
\newcommand{\Gcal}{\mathcal{G}} % subgoals
\newcommand{\Gopt}{\Gcal_{\textrm{opt}}} % subgoals used by the expert
\newcommand{\Acal}{\mathcal{A}} % primitive actions
\newcommand{\Mcal}{\mathcal{M}} % meta-controller class
\newcommand{\Dcal}{\mathcal{D}} % dataset
\newcommand{\Scal}{\mathcal{S}} % states
\newcommand{\hi}{\textsc{hi}\xspace}
\newcommand{\lo}{\textsc{lo}\xspace}
\newcommand{\fu}{\textsc{full}}
\newcommand{\chk}{\texttt{Inspect}}
\newcommand{\lbl}{\texttt{Label}}
\newcommand{\demo}{\texttt{HierDemo}}
\newcommand{\success}{\texttt{success}\xspace}
\newcommand{\terminal}{\texttt{terminal}\xspace}
\newcommand{\pseudo}{\texttt{pseudo}\xspace}%{\texttt{pseudo-reward}}
\newcommand{\True}{\textit{True}}
\newcommand{\False}{\textit{False}}

\newcommand{\pigopt}{\pi_g^\star}
\newcommand{\muopt}{\mu^\star}
\newcommand{\pifu}{\pi_\fu}
\newcommand{\pifuopt}{\pi_\fu^\star}
\newcommand{\Hhi}{H_\hi}
\newcommand{\Hlo}{H_\lo}
\newcommand{\Hfu}{H_\fu}
\newcommand{\Dg}{\Dcal_g}
\newcommand{\Dhi}{\Dcal_\hi}
\newcommand{\Pilo}{\Pi_\lo}
\newcommand{\Pifu}{\Pi_\fu}
\newcommand{\lbllo}{\lbl_\lo}
\newcommand{\chklo}{\chk_\lo}
\newcommand{\lblhi}{\lbl_\hi}
\newcommand{\chkfu}{\chk_\fu}
\newcommand{\lblfu}{\lbl_\fu}
\newcommand{\CLlo}{C_\lo^\textit{L}}
\newcommand{\CIlo}{C_\lo^\textit{I}}
\newcommand{\CLhi}{C_\hi^\textit{L}}
\newcommand{\CIfu}{C_\fu^\textit{I}}
\newcommand{\CLfu}{C_\fu^\textit{L}}
\newcommand{\rnd}{\xi}
\newcommand{\hh}{{h_\hi}} % time index at high level
\newcommand{\sh}{z} % new name for input of the meta-controller (which is a state)
\newcommand{\timepen}{\kappa}

\newcommand{\set}[1]{\{#1\}}

\newcommand{\bigParens}[1]{\bigl(#1\bigr)}
\newcommand{\BigParens}[1]{\Bigl(#1\Bigr)}

\newcommand{\Abs}[1]{\left\lvert#1\right\rvert}

\newcommand{\Fig}[1]{Figure~\ref{fig:#1}}

\newcommand{\mutil}{\tilde \mu} % maximizer \mu of surrogate return
\newcommand{\pigtil}{\tilde \pi_g} % maximizer \pi_g of surrogate return
\newcommand{\pigall}{\{\pigtil\}}
\newcommand{\rhopi}{\tilde \rho(\mu, \{\pi_g\})} 

\icmltitlerunning{Hierarchical Imitation and Reinforcement Learning}

\begin{document}

\twocolumn[
\icmltitle{Hierarchical Imitation and Reinforcement Learning}

% It is OKAY to include author information, even for blind
% submissions: the style file will automatically remove it for you
% unless you've provided the [accepted] option to the icml2017
% package.

% list of affiliations. the first argument should be a (short)
% identifier you will use later to specify author affiliations
% Academic affiliations should list Department, University, City, Region, Country
% Industry affiliations should list Company, City, Region, Country

% you can specify symbols, otherwise they are numbered in order
% ideally, you should not use this facility. affiliations will be numbered
% in order of appearance and this is the preferred way.
\icmlsetsymbol{equal}{*}

\begin{icmlauthorlist}
\icmlauthor{Hoang M. Le}{caltech}
\icmlauthor{Nan Jiang}{microsoft}
\icmlauthor{Alekh Agarwal}{microsoft}
\icmlauthor{Miroslav Dud\'ik}{microsoft}
\icmlauthor{Yisong Yue}{caltech}
\icmlauthor{Hal Daum\'e III}{maryland,microsoft}
\end{icmlauthorlist}

\icmlaffiliation{caltech}{California Institute of Technology, Pasadena, CA}
\icmlaffiliation{microsoft}{Microsoft Research, New York, NY}
\icmlaffiliation{maryland}{University of Maryland, College Park, MD}

\icmlcorrespondingauthor{Hoang M. Le}{hmle@caltech.edu}

% You may provide any keywords that you
% find helpful for describing your paper; these are used to populate
% the "keywords" metadata in the PDF but will not be shown in the document
\icmlkeywords{imitation learning, hierarchical reinforcement learning}

\vskip 0.3in
]

% this must go after the closing bracket ] following \twocolumn[ ...

% This command actually creates the footnote in the first column
% listing the affiliations and the copyright notice.
% The command takes one argument, which is text to display at the start of the footnote.
% The \icmlEqualContribution command is standard text for equal contribution.
% Remove it (just {}) if you do not need this facility.

\printAffiliationsAndNotice{}  % leave blank if no need to mention equal contribution
%\printAffiliationsAndNotice{\icmlEqualContribution} % otherwise use the standard text.

\begin{abstract}
We study how to effectively leverage expert feedback to learn sequential decision-making policies. We focus on problems with sparse rewards and long time horizons, which typically pose significant challenges in reinforcement learning. We propose an algorithmic framework, called \emph{hierarchical guidance}, that leverages the hierarchical structure of the underlying problem to integrate different modes of expert interaction. Our framework can incorporate different combinations of imitation learning (IL) and reinforcement learning (RL) at different levels, leading to dramatic reductions in both expert effort and cost of exploration. Using long-horizon benchmarks, including Montezuma's Revenge, we  demonstrate that our approach can learn significantly faster than hierarchical RL, and be significantly more label-efficient than standard IL. We also theoretically analyze labeling cost for certain instantiations of our framework.
\end{abstract}

\section{Introduction} \label{sec:intro} %% Story:
%%  - Long horizons -> use imitation learning
%%  - This paper: how to use expert feedback effectively
%%  - Approach: hierarchical learning
%%  - Contributions
%%     - Formalize the hierarchical IL
%%     - Two algorithms: IL / IL; IL / RL
%%     - Theory: benefits of the hierarchy
%%     - Experiments

%% Related work:
%%  - IL
%%  - Hierarchical RL
%%     - Feudal, hierarchical Q
%%  - Hierarchical modeling
%%     - robocup
%%  - Curriculum learning
%%  - Planning / learning with options
%%  - Option learning / discovery
%%  - (non-hierarchical) hybrids of IL + RL (addIL demonstrations into an experience buffer)
%%  - IL with different forms of feedback (with different costs?)

%\mdcomment{TODO: Include relevant references throughout introduction. Possibly also distill the most relevant additional related work into a paragraph or two.}

Learning good agent behavior from reward signals alone---the goal of reinforcement learning (RL)---is particularly difficult when the planning horizon is long and rewards are sparse.
One successful method for dealing with such long horizons is imitation learning (IL) \cite{abbeel2004apprenticeship,daume2009search,ross2011reduction,ho2016generative},  in which the agent learns by watching and possibly querying an expert. One limitation of existing imitation learning approaches is that they may require a large amount of demonstration data in long-horizon problems.

%   in which the agent learns not from reinforcement but by watching, and possibly querying, an expert. One limitation of existing imitation learning approaches is that they can require a large amount of demonstration data as the time horizon increases.

The central question we address in this paper is: \emph{when experts are available, how can we most effectively leverage their feedback?}
A common strategy to improve sample efficiency in RL over long time horizons is to exploit hierarchical structure of the problem~\cite{sutton1998intra,sutton1999between,kulkarni2016hierarchical,dayan1993feudal,vezhnevets2017feudal,dietterich2000hierarchical}.
Our approach leverages hierarchical structure in imitation learning. We study the case where the underlying problem is hierarchical, and subtasks can be easily elicited from an expert.
%, but has thus far not incorporated expert feedback into the learning process. \njcomment{Does the policy sketch paper count?}
%
\mdedit{Our key design principle is an algorithmic framework called \emph{hierarchical guidance}, in which feedback (labels) from the high-level expert is used to focus (guide) the low-level learner. The high-level expert ensures that low-level learning only occurs when necessary (when subtasks have not been mastered) and only over relevant parts of the state space. This differs from a na\"ive hierarchical approach which merely gives a subtask decomposition. Focusing on relevant parts of the state space speeds up learning (improves sample efficiency), while omitting feedback on the already mastered subtasks reduces expert effort (improves label efficiency).}

We begin by formalizing the problem of hierarchical imitation learning (\autoref{sec:formalism}) and carefully \mdedit{separate out} cost structures that naturally arise when the expert provides feedback at multiple levels of abstraction. \mdedit{We first apply hierarchical guidance to IL, derive hierarchically guided variants of behavior cloning and DAgger~\citep{ross2011reduction}, and theoretically analyze the benefits  (\autoref{sec:ilil}).
We next apply hierarchical guidance to the hybrid setting with high-level IL  and low-level RL~(\autoref{sec:ilrl}). This architecture is particularly suitable in settings where we have access to high-level semantic knowledge, the subtask horizon is sufficiently short, but the low-level expert is too costly or unavailable}.
%This leads to a\edit{n} \sout{novel} algorithm for combining imitation learning on top of reinforcement learning.
We demonstrate the efficacy of our approaches on a simple but extremely challenging maze domain, and on Montezuma's Revenge
(\autoref{sec:experiments}). \edit{Our experiments show that incorporating a modest amount of expert feedback can lead to dramatic improvements in performance compared to pure hierarchical RL.}\footnote{Code and experimental setups are available at \url{https://sites.google.com/view/hierarchical-il-rl}}

\sout{In the case where no expert feedback is available during training, our framework reduces to a standard form of hierarchical reinforcement learning \mbox{\cite{kulkarni2016hierarchical}}. We show in our experiments that incorporating a modest amount of expert feedback can lead to dramatic improvements in performance compared to conventional hierarchical RL.}

%%% Local Variables:
%%% mode: latex
%%% TeX-master: "icml_paper.tex"
%%% End:

\section{Related Work} \label{sec:related} %\mdcomment{TODO: (1) Include the MIT paper. (2) If this section stays in the main paper, then all the different paragraphs need to be connected better---right now it is a bit difficult to follow the thread. (3) Move to a later point in the paper: somewhere around conclusion/discussion or possibly into appendix.}

\edit{For brevity, we provide here a short overview of related work, and defer to Appendix~\ref{sec:others} for additional discussion.}

%Roughly speaking, related research to our work span two principle dimensions.   The first is learning hierarchical policy classes. The second is combining imitation learning with reinforcement learning.
\textbf{Imitation Learning.}
One can broadly dichotomize IL into passive collection of demonstrations (behavioral cloning) versus active collection of demonstrations. The former setting \citep{abbeel2004apprenticeship,ziebart2008maximum,syed2008game,ho2016generative} assumes that demonstrations are collected a priori and the goal of IL is to find a policy that mimics the demonstrations.  The latter setting \citep{daume2009search,ross2011reduction,ross2014reinforcement,chang2015learning,sun2017deeply} assumes an interactive expert that provides demonstrations in response to actions taken by the current policy. We explore extension of both approaches into hierarchical settings.

\textbf{Hierarchical Reinforcement Learning.}
Several RL approaches to learning hierarchical policies have been explored, foremost among them the options framework \citep{sutton1998intra,sutton1999between,fruit2017exploration}. %Options can be viewed as macro-actions that last for multiple time steps and accomplish subgoals.
It is often assumed that a useful set of options are fully defined a priori, and (semi-Markov) planning and learning only occurs at the higher level. In comparison, our agent does not have direct access to policies that accomplish such subgoals and has to learn them via expert or reinforcement feedback.
The closest hierarchical RL work to ours is that of \citet{kulkarni2016hierarchical}, which \mdedit{uses a similar hierarchical structure,
but no high-level expert and hence no hierarchical guidance}.
%\sout{As mentioned in the introduction, }\edit{T}heir approach can be viewed as a special case of our framework where no expert feedback is available during training.

%Methodologically, perhaps the closest related work is \citep{andreas2016modular}, which address the multi-task reinforcement learning problem by assuming access to symbolic descriptions of subgoals, without knowing what those symbols mean or how to execute them. In the multi-task case, common subgoals form the shared structured and thus form the basis for learning and reusing subgoals from one task to the next. Learning is done via policy gradient method, with careful choice of baseline functions to ensure efficient learning.

\textbf{Combining Reinforcement and Imitation Learning.} The idea of combining IL and RL is not new \cite{nair2018overcoming,hester2018deep}.  However, previous work focuses on flat policy classes that use IL as a ``pre-training'' step (e.g., by pre-populating the replay buffer with demonstrations). In contrast, we consider feedback at multiple levels for a hierarchical policy class, with different levels potentially receiving different types of feedback (i.e., imitation at one level and reinforcement at the other). Somewhat related to our hierarchical expert supervision is the approach of~\citet{andreas2016modular}, which assumes access to symbolic descriptions of subgoals, without knowing what those symbols mean or how to execute them. Previous literature has not focused much on comparisons of sample complexity between IL and RL, with the exception of the recent work of~\citet{sun2017deeply}.

\section{Hierarchical Formalism} \label{sec:formalism} %Levels: Hi, Lo etc.

For simplicity, we consider environments with a natural two-level hierarchy; the $\hi$ level corresponds to choosing subtasks, and the $\lo$ level corresponds to executing those subtasks.
For instance, an agent's overall goal may be to leave a building. At the $\hi$ level, the agent may first choose the subtask \emph{``go to the elevator,''} then \emph{``take the elevator down,''} and finally \emph{``walk out.''} Each of these subtasks needs to be executed at the $\lo$ level by actually navigating the environment, pressing buttons on the elevator, etc.\footnote{\edit{An important real-world application is in goal-oriented dialogue systems. For instance, a chatbot assisting a user with reservation and booking for flights and hotels \cite{peng2017composite, el2017frames} needs to navigate through multiple turns of conversation. The chatbot developer designs the hierarchy of subtasks, such as \emph{ask\_user$\!\_\,$goal}, \emph{ask\_dates}, \emph{offer$\!\_\,$flights}, \emph{confirm}, etc.
%\emph{greet}, \emph{inform}, \emph{offer}, \emph{request}, \emph{compare\_price}, \emph{confirm}, \emph{more\_info}, etc.
Each subtask consists of several turns of conversation. Typically a global state tracker exists alongside the hierarchical dialogue policy to ensure that cross-subtask constraints are satisfied.}}

Subtasks, which we also call \emph{subgoals}, are denoted as $g \in \Gcal$,
and the primitive actions are denoted as $a \in \Acal$.
An agent (also referred to as learner) acts by iteratively choosing a subgoal $g$, carrying it out by executing a sequence of actions $a$ until completion, and then picking a new subgoal. The agent's choices can depend on an observed
state $s\in\Scal$.\footnote{While we use the term state for simplicity, we do not require the environment to be fully observable or Markovian.}
%our approach does not necessarily require $s$ to be Markovian, especially when we have policy learning algorithms at both levels.}
%
%\mdcomment{Should we make a footnote to clarify that the environment is not necessarily fully observable and that any possible reward would be encapsulated as part of the state?}
%
We assume that the horizon at the $\hi$ level is $\Hhi$, i.e., a trajectory uses at most $\Hhi$ subgoals, and the horizon at the $\lo$ level is $\Hlo$, i.e., after at most $\Hlo$ primitive actions, the agent either accomplishes the subgoal or needs to decide on a new subgoal.
The total number of primitive actions in a trajectory is thus at most $\Hfu\coloneqq\Hhi\Hlo$.

The hierarchical learning problem is to simultaneously
learn a \hi-level policy $\mu:\Scal\to\Gcal$, called the
\emph{meta-controller}, as well as the subgoal policies
${\pi_g:\Scal\to\Acal}$ for each ${g\in\Gcal}$, called \emph{subpolicies}. The aim of the learner is to achieve a high reward when its meta-controller and subpolicies are run together.
For each subgoal~$g$, we also have a (possibly learned) termination function~$\beta_g:\Scal\to\set{\True,\False}$, which terminates the execution of $\pi_g$.
%$\mu \in \Mcal$, where $\Mcal$ is a policy class at the \hi-level.
%$\pi_g \in \Pi_g$, where $\Pi_g$ is a \lo-level policy class.
%Finally, we will denote by $\mu^\star$ and $\pi_g^\star$, the optimal meta-controller policy and subgoal policies for each $g$.
The hierarchical agent behaves as follows:

\begin{algorithmic}[1]
  \FOR{$h_\hi = 1 \dots \Hhi$}
  \STATE observe state $s$ and choose subgoal $g \gets \mu(s)$
  \FOR{$h_\lo = 1 \dots \Hlo$}
  \STATE observe state $s$
  \STATE \textbf{if} $\beta_g(s)$ \textbf{then break}
  \STATE choose action $a \gets \pi_g(s)$
  \ENDFOR
  \ENDFOR
\end{algorithmic}

%\mdcomment{I'm intentionally avoiding double indexing and double subscripts below---we definitely don't want to write
%monstrosities such as $s_{h_\hi,h_\lo}$. I hope things are still clear. If we truly need two indices, I'm in favor
%of replacing $h_\hi$ with $h$, and $h_\lo$ with $\ell$, but then we should also replace $\Hhi$ with $H$ and $\Hlo$ with
%$L$, and perhaps we don't need a special symbol for the full trajectory length, which would become $HL$.}
%\njcomment{The issue with $H$ and $L$ is that $\CLlo$ uses $L$ in the superscript for ``label''.}

The execution of each subpolicy $\pi_g$ generates a \emph{\lo-level trajectory} $\tau=(s_1,a_1,\dotsc,s_H,a_H,s_{H+1})$ with $H\le\Hlo$.\footnote{%
   The trajectory might optionally include a reward signal after each primitive action, which might either come from the environment, or be a pseudo-reward as we will see in Section~\ref{sec:ilrl}.%
}
%and when $H<\Hlo$ then $s_{H+1}$ is the terminating state with $\beta_g(s_{H+1})=\True$.
The overall behavior results in a \emph{hierarchical trajectory} $\sigma=(s_1,g_1,\tau_1,s_2,g_2,\tau_2,\dotsc)$,
where the last state of each \lo-level trajectory $\tau_h$
coincides with the next state $s_{h+1}$ in $\sigma$ and the first state of the next \lo-level trajectory $\tau_{h+1}$.
%We use $\tau_\hi\coloneqq(s_1,g_1,s_2,g_2,\dotsc)$ to denote the \emph{\hi-level trajectory}, which is essentially $\sigma$ without the \lo-level trajectories $\tau_h$.
The subsequence of $\sigma$ which excludes the \lo-level trajectories $\tau_h$ is called the
\emph{\hi-level trajectory}, $\tau_\hi\coloneqq(s_1,g_1,s_2,g_2,\dotsc)$.
%%
%% MD: For the moment, I'm going back to the original phrasing, because it is important to consider hi-level trajectory in the context of some hierarchical execution,
%%     which was de-emphasized in the rewrite.
%%
Finally, the \emph{full trajectory}, $\tau_\fu$, is the concatenation of all the \lo-level trajectories.

%(if $h<\Hhi$).
%\hlecomment{It seems that HierDemo is nothing more than $\lblhi$ + $\lbllo$, and only plays a role in shortening the pseudo code for hierarchical behavioral cloning. Should we bring it up here? }

%\mdcomment{\demo is a bit different, because it does not work by labeling the agent trajectory, but instead produces the expert trajectory from scratch. I also wanted to call it out as first,
%because it is in some way the most natural form of supervision. That said, I'm happy to drop}

We assume access to an \emph{expert}, endowed with a meta-controller $\mu^\star$, subpolicies $\pi^\star_g$, and termination functions $\beta^\star_g$,
who can provide one or several types of supervision:
\vspace{-0.07in}
% at training time:
%
\begin{itemize}
%\item Subgoal termination: the expert can say when the agent has completed its current subgoal; we denote this function as $\beta$.
\item $\demo(s)$: \emph{hierarchical demonstration}. The expert executes its hierarchical policy starting from $s$ and returns the resulting hierarchical trajectory $\sigma^\star = (s^\star_1,g^\star_1,\tau^\star_1,s^\star_2,g^\star_2,\tau^\star_2,\dotsc)$, where $s^\star_1=s$.
\vspace{-0.05in}
\item $\lblhi(\tau_\hi)$: \emph{\hi-level labeling}. The expert provides a good next subgoal
     at each state of a given \hi-level trajectory $\tau_\hi=(s_1,g_1,s_2,g_2,\dotsc)$, %so if $\tau_\hi$ is of length $H$,
     yielding a labeled data set $\set{(s_1,g^\star_1),(s_2,g^\star_2),\dotsc}$.
     %which results in a list of $|\tau_\hi|$ state-subgoal pairs.
\vspace{-0.05in}
\item $\lbllo(\tau;g)$: \emph{\lo-level labeling}. The expert provides a good next primitive action towards a given subgoal~$g$
     at each state of a given \lo-level trajectory $\tau=(s_1,a_1,s_2,a_2,\dotsc)$,
     yielding a labeled data set $\set{(s_1,a^\star_1),(s_2,a^\star_2),\dotsc}$.
     %which results in a list of $|\tau|$ state-action pairs.
\vspace{-0.05in}
\item $\chklo(\tau;g)$: \emph{\lo-level inspection}. Instead of annotating every state of a trajectory with a good action, the expert only verifies whether a subgoal $g$ was accomplished, returning either \textit{Pass} or \textit{Fail}.
\vspace{-0.05in}
\item $\lblfu(\tau_\fu)$: \emph{full labeling}. The expert labels the agent's full trajectory $\tau_\fu=(s_1,a_1,s_2,a_2,\dotsc)$,
     from start to finish, ignoring hierarchical structure, yielding a labeled data set $\set{(s_1,a^\star_1),(s_2,a^\star_2),\dotsc}$.
\vspace{-0.05in}
\item $\chkfu(\tau_\fu)$: \emph{full inspection}. The expert verifies whether the agent's overall goal was accomplished, returning either \textit{Pass} or \textit{Fail}.
\end{itemize}
\vspace{-0.05in}

When the agent learns not only the subpolicies $\pi_g$, but also termination functions $\beta_g$, then $\lbllo$ also returns good termination values $\omega^\star\in\set{\True,\False}$ for
each state of $\tau=(s_1,a_1\dotsc)$, yielding a data set $\set{(s_1,a^\star_1,\omega^\star_1),\dotsc}$.

Although $\demo$ and $\lbl$ can be both generated by the expert's hierarchical policy $(\mu^\star,\set{\pi^\star_g})$, they differ in the mode of expert interaction.
$\demo$ returns a hierarchical trajectory \emph{executed by the expert}, as required for passive IL, and  enables a hierarchical version of behavioral cloning \citep{abbeel2004apprenticeship,syed2008game}.  $\lbl$ operations provide labels \textit{with respect to the learning agent's trajectories}, as required for interactive IL. $\lblfu$ is the standard query used in prior work on learning flat policies~\citep{daume2009search,ross2011reduction}, and $\lblhi$ and $\lbllo$ are its hierarchical extensions.

$\chk$ operations are newly introduced in this paper, and form a cornerstone of \mdedit{our interactive hierarchical guidance} protocol that enables substantial savings in label efficiency.  They can be viewed as ``lazy'' versions of the corresponding $\lbl$ operations,  requiring less effort.
Our underlying assumption is that if the given hierarchical trajectory $\sigma=\set{(s_h,g_h,\tau_h)}$ agrees with the expert on \hi level,
i.e., $g_h=\mu^\star(s_h)$, and \lo-level trajectories pass the inspection, i.e., $\chklo(\tau_h;g_h)=\textit{Pass}$, then the resulting full trajectory must also pass the full inspection, $\chkfu(\tau_{\fu})=\textit{Pass}$.
This means that a hierarchical policy need not always agree with the expert's execution at \lo level to succeed in the overall task.

%\mdcomment{Maybe this is a good place to introduce expert meta-controller and subpolicies $\mu^\star$ and $\pi_g^\star$, and describe a bit more formally how they give rise to $\demo$ the three $\lbl$ operations.
%Also, we need to say something slightly
%formal about how $\chk$ operations relate to $\mu^\star$ and $\pi_g^\star$. For instance, maybe we should posit that
%(1) following expert policies always results in passing $\chklo$ and $\chkfu$; (2) following $\mu^\star$ while executing $\pi_g$ such that every \lo-level trajectory passes $\chklo$ results in passing $\chkfu$.}

Besides algorithmic reasons, the motivation for separating the types of feedback is that different expert queries will typically require different amount of effort, which we refer to as \emph{cost}.
We assume the costs of the $\lbl$ operations are $\CLhi$, $\CLlo$ and $\CLfu$, the costs of each $\chk$ operation are $\CIlo$ and $\CIfu$.
%assuming that the \lo-level costs are the same across all subgoals~$g$.
%(where, for convenience, we assume that the cost is the same for all subgoals $g$).
In many settings, \lo-level inspection will require significantly less effort than \lo-level labeling,
i.e., $\CIlo \ll \CLlo$. For instance, identifying if a robot has successfully navigated to the elevator is presumably much easier than labeling an entire path to the elevator.
%Similarly, in most cases we expect $\CIfu\ll\CLfu$.
One reasonable cost model, natural for the environments in our experiments, is to assume that $\chk$ operations take time $O(1)$ and work by checking the final state of
the trajectory, whereas $\lbl$ operations take time proportional to the trajectory length, which is $O(H_\hi)$, $O(H_\lo)$ and $O(H_\hi H_\lo)$ for our three $\lbl$ operations.

%%% Local Variables:
%%% mode: latex
%%% TeX-master: "icml_paper.tex"
%%% End:

\section{Hierarchically Guided Imitation Learning} \label{sec:ilil} %% - Algorithm
%% - Theory
%%    - Present mistake bound, contrast with flat IL, discuss costs
%%    - Present regret bound(s)

%\mdcomment{TODO: algorithms are done; the text needs to be written; Table of notation updated or dropped.}

%%%%%%%%%%%%%%%%%%%%%%%%%%%%%%%

\mdedit{%
\emph{Hierarchical guidance} is an algorithmic design principle in which the feedback from high-level expert guides the low-level learner in two different ways: (i) the high-level expert ensures that low-level expert is only queried when necessary (when the subtasks have not been mastered yet), and (ii) low-level learning is limited to the relevant parts of the state space.
We instantiate this framework first within passive learning from demonstrations, obtaining \emph{hierarchical behavioral cloning} (Algorithm~\ref{alg:hierarchical_bc}),
%a simple extension of the flat (non-hierarchical) behavior cloning
and then within interactive imitation learning, obtaining \emph{hierarchically guided DAgger} (Algorithm~\ref{alg:hierarchical_dagger}), our best-performing algorithm.}
%and provide theoretical analysis of its cost efficiency compared with the flat (non-hierarchical) approach.
%The algorithm uses hierarchical behavioral cloning far a warm start, but then switches to the interactive mode of expert labeling.}

%In this section we introduce our new algorithm, hierarchical DAgger (Algorithm~\ref{alg:hierarchical_dagger}), and provide theoretical analysis of its cost efficiency compared with a flat approach. The algorithm uses hierarchical behavior cloning (Algorithm~\ref{alg:hierarchical_bc}) as a warm-start procedure, which by itself is also a hierarchical imitation learning algorithm and only requires passive demonstrations.

\begin{algorithm}[t]
   \caption{Hierarchical Behavioral Cloning (\textbf{h-BC})}
   \label{alg:hierarchical_bc}
   \begin{small}
\begin{algorithmic}[1]
   \STATE Initialize data buffers $\Dhi \leftarrow \emptyset$ and $\Dg \leftarrow \emptyset$, $g\in\Gcal$
   \FOR{$t=1,\ldots,T$}
       \STATE Get a new environment instance with start state $s$
       \STATE $\sigma^\star\gets\demo(s)$
       \FORALL{$(s^\star_h,g^\star_h,\tau^\star_h)\in\sigma^\star$}
          \STATE Append $\Dcal_{g^\star_h}\gets\Dcal_{g^\star_h}\cup\,\tau^\star_h$
          \STATE Append $\Dhi\gets\Dhi \cup\set{(s^\star_h,g^\star_h)}$
       \ENDFOR
   \ENDFOR
   \STATE Train subpolicies $\pi_g \leftarrow \texttt{Train}(\pi_g,\Dg)$ for all $g$
   \STATE Train meta-controller $\mu \leftarrow \texttt{Train}(\mu,\Dhi)$
\end{algorithmic}
   \end{small}
\end{algorithm}

\subsection{Hierarchical Behavioral Cloning (h-BC)}
\label{sec:hierarchical_bc}

%\mdcomment{This is the simplest approach, describe very briefly; give examples of what \textit{Train} might be.}

%Typically, the flat behavior cloning procedure requires the expert to design full demonstration from scratch
We consider a natural extension of behavioral cloning to the hierarchical setting (Algorithm~\ref{alg:hierarchical_bc}). The expert provides a set of  hierarchical demonstrations $\sigma^\star$, each consisting of \lo-level trajectories $\tau^\star_h = \set{(s^\star_\ell,a^\star_\ell)}_{\ell=1}^{H_\lo}$ as well as a \hi-level trajectory $\tau_\hi^\star = \set{(s^\star_h,g^\star_h)}_{h=1}^{H_\hi}$. %This signifies the expert's hierarchical decomposition.
%the flat behavior cloning procedure simply consists of the operation $\lblfu(\tau_\fu)$, which may overlap with $\lbllo(\tau;g)$. Beside giving \lo-level trajectories $\tau_h$, hierarchical demonstration also provides \hi-level labeling $\lblhi(\tau_\hi) = \set{(s_h,g_h)}$, which signifies the expert's hierarchical decomposition.
We then run \texttt{Train} (lines 8--9) to find the subpolicies $\pi_g$ that best predict $a^\star_\ell$ from $s^\star_\ell$, and meta-controller $\mu$ that best predicts $g^\star_h$ from $s^\star_h$, respectively.
\texttt{Train} can generally be any supervised learning subroutine, such as stochastic optimization for neural networks or some batch training procedure. When termination functions $\beta_g$ need to be learned as part of the hierarchical policy, the labels $\omega^\star_g$ will be provided by the expert as part of $\tau^\star_h = \set{(s_\ell^\star,a^\star_\ell,\omega^\star_\ell)}$.\footnote{%
In our hierarchical imitation learning experiments, the termination functions are all learned. Formally, the termination signal $\omega_g$, can be viewed as part of an augmented action at $\lo$ level.}
\mdedit{In this setting, hierarchical guidance is automatic, because subpolicy demonstrations only occur in relevant parts of the state space}.\looseness=-1

\begin{algorithm}[t]
   \caption{Hierarchically Guided DAgger (\textbf{hg-DAgger})}
   \label{alg:hierarchical_dagger}
   \begin{small}
\begin{algorithmic}[1]
   \STATE Initialize data buffers $\Dhi \leftarrow \emptyset$ and $\Dg \leftarrow \emptyset$, $g\in\Gcal$
   \STATE Run Hierarchical Behavioral Cloning (Algorithm~\ref{alg:hierarchical_bc})\\
          \hspace{1em}up to $t=T_{\textrm{warm-start}}$
   \FOR{$t=T_{\textrm{warm-start}}+1,\ldots,T$}
   \STATE Get a new environment instance with start state $s$
   \STATE Initialize $\sigma\gets\emptyset$
   \REPEAT
      \STATE $g\gets\mu(s)$
      \STATE Execute $\pi_g$, obtain \lo-level trajectory $\tau$
      \STATE Append $(s,g,\tau)$ to $\sigma$
      \STATE $s\gets\text{the last state in $\tau$}$
   \UNTIL{end of episode}
   \STATE Extract $\tau_\fu$ and $\tau_\hi$ from $\sigma$
   \IF{$\chkfu(\tau_\fu)=\textit{Fail}$}
      \STATE $\Dcal^\star\gets\lblhi(\tau_\hi)$
      \STATE Process $(s_h,g_h,\tau_h)\in\sigma$ in sequence as long as\\
             $g_h$ agrees with the expert's choice $g^\star_h$ in $\Dcal^\star$:
      \begin{ALC@g}
         \IF{$\chk(\tau_h;g_h)=\textit{Fail}$}
            \STATE Append $\Dcal_{g_h}\gets\Dcal_{g_h}\cup\,\lbllo(\tau_h;g_h)$
            \STATE \textbf{break}
         \ENDIF
      \end{ALC@g}
      \STATE Append $\Dhi\gets\Dhi\cup\,\Dcal^\star$
   \ENDIF
   \STATE Update subpolicies $\pi_g \leftarrow \texttt{Train}(\pi_g,\Dg)$ for all $g$
   \STATE Update meta-controller $\mu \leftarrow \texttt{Train}(\mu,\Dhi)$
   \ENDFOR
\end{algorithmic}
\end{small}
\end{algorithm}

\subsection{\edit{Hierarchically Guided DAgger (hg-DAgger)}}
\label{sec:hierarchical_imitation}

%\mdcomment{Describe the algorithm. Maybe set up some of the intuitions
%that will be spelled out more precisely in the analysis:
%(1) we are only learning new subpolicies along good trajectories;
%(2) inspection operations allow us to ``jump ahead'' in labeling effort.}

\begin{comment}
%Algorithm \ref{alg:hierarchical_bc} leverages expert hierarchical feedback in a similar fashion to the flat behavior cloning.
%In hierarchical settings, $\demo$ can reasonably be viewed as the most natural form of supervision.
While Algorithm~\ref{alg:hierarchical_bc} leverages hierarchical expert feedback,
%unless the expert gives a diverse range of (hierarchical) demonstrations,
the well-known distribution mismatch problem between learning and execution can still occur when reducing sequential decision making to supervised learning. %such as $\texttt{Train}$
%\citep{daume2009search,ross2011reduction}. %This distribution mismatch is magnified when the states are only partially observed.
Interactive imitation learning algorithms, such as \edit{SEARN \cite{daume2009search} and} DAgger \cite{ross2011reduction}, address this issue by having the expert actively provide feedback \textit{on the learner's trajectories}. %, which we referred to as $\lblfu$.
%(and $\lblhi, \lbllo$ for hierarchical setting).
However, existing interactive algorithms cannot leverage hierarchical feedback, and typically invoke $\lblfu$ from start to finish. % executing the learner's policy at each training round and then invoking $\lblfu$ from start to finish.
\end{comment}

Passive IL, e.g., behavioral cloning, suffers from the distribution mismatch between the learning and execution distributions.
This mismatch is addressed by
interactive IL algorithms, such as SEARN \cite{daume2009search} and DAgger \cite{ross2011reduction}, where the expert provides
correct actions along the learner's trajectories through the operation $\lblfu$. A na\"ive hierarchical implementation would provide correct labels along
the entire hierarchical trajectory via $\lblhi$ and $\lbllo$. We next show how to use hierarchical guidance to decrease \lo-level expert costs.

\mdedit{%
We leverage two $\hi$-level query types: $\chklo$ and $\lblhi$. We use
$\chklo$ to verify whether the subtasks are successfully completed and $\lblhi$ to check whether we are staying in the relevant part of the state space. The details are presented in Algorithm~\ref{alg:hierarchical_dagger},
which uses DAgger as the learner on both levels, but the scheme can be adapted to other interactive imitation learners}.

In each episode, the learner executes the hierarchical policy, including choosing a subgoal (line~7), executing the \lo-level trajectories, i.e., rolling out the subpolicy $\pi_g$ for the chosen subgoal, and terminating the execution according to $\beta_g$ (line~8). Expert only provides feedback when the agent fails to execute the entire task, as verified by $\chkfu$ (line~13). When $\chkfu$ fails, the expert first labels the correct subgoals via $\lblhi$ (line 14), and only performs \lo-level labeling as long as the learner's meta-controller chooses the correct subgoal $g_h$ (line 15),  %attempts to execute a ``good'' subgoal,
but its subpolicy fails (i.e., when $\chklo$ on line 16 fails). \mdedit{Since all the preceding subgoals were chosen and executed correctly, and the current subgoal is also correct,
\lo-level learning is in the ``relevant'' part of the state space. However, since the subpolicy execution failed, its learning has not been mastered yet.
%This exactly corresponds to hierarchical guidance.
We next analyze the savings in expert cost that result from hierarchical guidance.\looseness=-1}

\textbf{Theoretical Analysis.}
We analyze the cost of hg-DAgger in comparison with flat DAgger under somewhat stylized assumptions.
We assume that the learner aims to learn the meta-controller~$\mu$ from some policy class~$\Mcal$, and subpolicies $\pi_g$ from some class $\Pilo$.
The classes $\Mcal$ and $\Pilo$ are finite (but possibly exponentially large) and the task is realizable, i.e., the expert's policies can be found in the corresponding classes: $\mu^\star \in \Mcal$, and $\pi_g^\star \in \Pilo$, $g\in\Gcal$. This allows us to use the \emph{halving algorithm}~\cite{shalev2012online} as the online learner on both levels.
(The implementation of our algorithm does not require these assumptions.)

%In order to calculate the total cost of the expert, we need to define the atomic cost associated with each type of expert operations:
%\begin{itemize}[nolistsep,noitemsep]
%\item $\CLlo$: cost of $\lbllo(\cdot)$.
%\item $\CIlo$: cost of $\chklo(\cdot)$.
%\item $\CLhi$: cost of $\lblhi(\cdot)$.
%\item $\CIfu$: cost of $\chkfu(\cdot)$.
%\item $\CLfu$: cost of $\lblfu(\cdot)$.
%\end{itemize}
%\begin{align} \label{eq:total_mistake}
%\sum_{t=1}^T \indi{\exists h, \textrm{s.t.~} \pifuopt \textrm{ and } \pi_\fu^t ~\textrm{disagrees on}~ a_h^t}.
%\end{align}

The halving algorithm maintains a version space over policies, acts by a majority decision,
and when it makes a mistake, it removes all the erring policies from the version space. In the hierarchical setting,
it therefore makes at most $\log\Abs{\Mcal}$ mistakes on the \hi level, and at most $\log\Abs{\Pilo}$ mistakes when learning each $\pi_g$.
The mistake bounds can be further used to upper bound the total expert cost in both \mdedit{hg-DAgger and flat DAgger. To enable an apples-to-apples comparison, we assume that the flat DAgger learns over the policy class $\Pifu = \{(\mu, \{\pi_g\}_{g\in\Gcal}):\:{\mu\in \Mcal}, {\pi_g \in \Pilo}\}$, but is otherwise oblivious to the hierarchical task structure.} The bounds depend on the cost of performing different types of operations, as defined at the end of Section~\ref{sec:formalism}. \mdedit{We consider a modified version of flat DAgger} that first calls $\chkfu$, and only requests labels ($\lblfu$) if the inspection fails.
The proofs are deferred to Appendix~\ref{sec:app_theory}.
\begin{theorem}
\label{thm:cost_hier}
Given finite classes $\Mcal$ and $\Pilo$ and realizable expert policies, the total cost incurred by the expert in hg-DAgger by round $T$ is bounded by
\begin{align}
\notag
T \CIfu
& + \bigParens{\log_2|\Mcal| + |\Gopt| \log_2|\Pilo|}(\CLhi + \Hhi \CIlo)
\\
\label{eq:cost_hier}
& + \bigParens{|\Gopt| \log_2|\Pilo|}\CLlo,
\end{align}
where $\Gopt\subseteq\Gcal$ is the set of the subgoals actually used
by the expert, $\Gopt\coloneqq\mu^\star(\Scal)$.
%is the set of subgoals that can be considered as \texttt{good} by the expert (see Line 12 in Algorithm~\ref{alg:hierarchical_dagger}). %that will be invoked with non-zero probability when rolling out episodes according to $\pifuopt$.
\end{theorem}

\begin{theorem}
\label{thm:cost_flat}
Given the full policy class $\Pifu = \{(\mu, \{\pi_g\}_{g\in\Gcal}):\:{\mu\in \Mcal}, {\pi_g \in \Pilo}\}$ and a realizable expert policy, the total cost incurred by the expert in
flat DAgger by round $T$ is bounded by
\begin{align}
\label{eq:cost_flat}
T \CIfu  + \bigParens{\log_2|\Mcal| + |\Gcal| \log_2|\Pilo|} \CLfu.
\end{align}
\end{theorem}

Both bounds have the same leading term, $T\CIfu$, the cost of full inspection, which is incurred every round and can be viewed as the ``cost of monitoring.''
In contrast, the remaining terms can be viewed as the ``cost of learning'' in the two settings, and include terms coming
from their respective mistake bounds. The ratio of the cost of hierarchically guided learning to the flat learning is then bounded as
\begin{align}
\label{eq:cost_ratio}
\frac{\text{Eq.~\eqref{eq:cost_hier}} - T \CIfu}{\text{Eq.~\eqref{eq:cost_flat}} - T \CIfu}
\le \frac{\CLhi + \Hhi \CIlo + \CLlo}{\CLfu},
\end{align}
where we applied the upper bound $|\Gopt| \le |\Gcal|$. The savings thanks to hierarchical guidance depend on the specific costs. Typically, we expect the inspection costs to be $O(1)$,
if it suffices to check the final state, whereas labeling costs scale linearly with the length of the trajectory. The cost ratio is then $\propto\frac{\Hhi+\Hlo}{\Hhi\Hlo}$. Thus,
we realize most significant savings if the horizons on each individual level are substantially shorter than the overall horizon. In particular, if $\Hhi=\Hlo=\sqrt{\Hfu}$, the hierarchically guided
approach reduces the overall labeling cost by a factor of $\sqrt{\Hfu}$. More generally, whenever $\Hfu$ is large, we reduce the costs of learning be at least a constant factor---a significant gain if this is a saving in the effort of a domain expert.

%Of course, the values of different type of costs depend on individual applications, which determines the ratio in Eq.\eqref{eq:cost_ratio}. However, in a typical situation where inspection cost is substantially lower than labelling cost, the dominating part of the numerator corresponds to $\CLhi + \CLlo$, that is, to provide $\Hhi$ $\hi$-level labels (subgoals) and $\Hlo$ $\lo$-level labels (actions). Comparing to the denominator which corresponds to providing $\Hhi \Hlo$ $\lo$-level labels, we can see that the hierarchical approach yields significant saving in expert cost when checking subgoal attainment is easy and $\hi$-level labels are not significantly more expensive than $\lo$-level labels.
%%The RHS is less than one when the cost of labelling the primitive actions of a whole trajectory exceeds the cost of first labelling and inspecting subgoals, and then labelling the primitive actions of a single subgoal.

%%% Local Variables:
%%% mode: latex
%%% TeX-master: "icml_paper.tex"
%%% End:

\section{Hierarchically Guided IL\,/\,RL} \label{sec:ilrl} \begin{algorithm}[tb]
	\caption{Hierarchically Guided DAgger\,/\,$Q$-learning\\\hphantom{\textbf{Algorithm 3}} (\textbf{hg-DAgger/Q})}
	\label{alg:hybrid_il_rl}
   \begin{small}
	\begin{algorithmic}[1]
%		\INPUT $\terminal(s;g)$ and $\success(s;g)$ indicating
%		termination and successful completion of $g$ in $s$
		\INPUT Function $\pseudo(s;g)$ providing the pseudo-reward
% in state $s$ when executing subgoal $g$
        \INPUT Predicate $\terminal(s;g)$ indicating the termination of~$g$
%$\terminal(s;g)$ and $\pseudo(s;g)$ indicating
%termination $g$ in $s$ and pseudo-reward while in $s$ and $g$
		\INPUT Annealed exploration probabilities $\epsilon_g>0$, $g\in\Gcal$
		%\INPUT Per-step penalty in pseudo-reward $\timepen>0$
		\STATE Initialize data buffers $\Dhi \leftarrow \emptyset$ and $\Dg \leftarrow \emptyset$, $g\in\Gcal$
		\STATE Initialize subgoal $Q$-functions $Q_g$, $g\in\Gcal$
		\FOR{$t=1,\ldots,T$}
		\STATE Get a new environment instance with start state $s$
        \STATE Initialize $\sigma\gets\emptyset$
		\REPEAT
		\STATE $s_\hi\gets s,\,g\gets\mu(s)$ and initialize $\tau\gets\emptyset$
		\REPEAT
		\STATE $a\gets\epsilon_g\text{-greedy}(Q_g,s)$
		%		\STATE $a\gets
		%		\begin{cases}
		%		\argmax_{a'} Q_g(s,a')
		%		&
		%		\text{w.p.\ $1-\epsilon_g$}
		%		\\
		%		\text{uniformly random}
		%		&
		%		\text{otherwise}
		%		\end{cases}$
		\STATE Execute $a,\,\text{next state $\tilde{s}$},\,\tilde{r}\gets \pseudo(\tilde{s};g)$
%		\STATE $\tilde{r}\gets 1 $ if $\success(\tilde{s};g)$
		%		\STATE $\tilde{r}\gets
		%		\begin{cases}
		%		1&\text{if $\success(\tilde{s};g)$}
		%		\\
		%		-\timepen&\text{otherwise}
		%		\end{cases}$
		\STATE Update $Q_g$:
		~a (stochastic) gradient descent step\\
		~\hphantom{$Q_g$-update:} on a minibatch from $\Dg$
		\STATE Append $(s,a,\tilde{r},\tilde{s})$ to $\tau$ and update $s\gets\tilde{s}$
		\UNTIL{$\terminal(s;g)$}
		\STATE Append $(s_\hi,g,\tau)$ to $\sigma$
		\UNTIL{end of episode}
		\STATE Extract $\tau_\fu$ and $\tau_\hi$ from $\sigma$
		\IF{$\chkfu(\tau_\fu)=\textit{Fail}$}
		\STATE $\Dcal^\star\gets\lblhi(\tau_\hi)$
		\STATE Process $(s_h,g_h,\tau_h)\in\sigma$ in sequence as long as\\
		$g_h$ agrees with the expert's choice $g^\star_h$ in $\Dcal^\star$:
		\begin{ALC@g}
			\STATE Append $\Dcal_{g_h}\gets\Dcal_{g_h}\cup\,\tau_h$
		\end{ALC@g}
		Append $\Dhi\gets\Dhi\cup\,\Dcal^\star$
		\ELSE
		\STATE Append $\Dcal_{g_h}\gets\Dcal_{g_h}\cup\,\tau_h$ for all $(s_h,g_h,\tau_h)\in\sigma$
		\ENDIF
		\STATE Update meta-controller $\mu \leftarrow \texttt{Train}(\mu,\Dhi)$
		\ENDFOR
	\end{algorithmic}
   \end{small}
\end{algorithm}

\begin{comment}
\edit{hg-IL is based on} \sout{Hierarchical DAgger was motivated by} the idea that it is generally easier for expert to teach learning agent at the \hi level instead of supervising at the \lo level. We further carry this idea to the reinforcement learning setting, where we let the agent learn the subpolicies from reinforcement signal alone. \edit{Our Hierarchically Guided Imitation-Reinforcement Learning hybrid (\textbf{hg-IL/RL}) relies on expert knowledge to specify subgoals and hence, the hierarchical decomposition.}\footnote{\edit{This is consistent with several prior hierarchical RL methods, such as the options framework \cite{sutton1999between}, MAXQ decomposition \cite{dietterich2000hierarchical}, UVFA \cite{schaul2015universal} and h-DQN \cite{kulkarni2016hierarchical}.}}
While our approach allows any imitation learning at the \hi level and any reinforcement learning at the \lo level, for concreteness, we present the variant with DAgger and $Q$-learning in Algorithm~\ref{alg:hybrid_il_rl}.
%Algorithm~\ref{alg:hybrid_il_rl} provides a meta-algorithm that uses imitation learning  at the \hi level and reinforcement learning at the \lo level. We outline an instantiation with DAgger and $Q$-Learning in the pseudo-code.
\end{comment}

Hierarchical guidance also applies in the hybrid setting with interactive IL on the \hi level and RL on the \lo level.
The \hi-level expert provides the hierarchical decomposition, including the pseudo-reward function for each subgoal,\footnote{%
This is consistent with many hierarchical RL approaches, including options \cite{sutton1999between}, MAXQ \cite{dietterich2000hierarchical}, UVFA \cite{schaul2015universal} and h-DQN \cite{kulkarni2016hierarchical}.}
and is also able to pick a correct subgoal at each step. Similar to hg-DAgger, the labels from \hi-level expert are used not only to train the meta-controller $\mu$, but also to
limit the \lo-level learning to the relevant part of the state space. In Algorithm~\ref{alg:hybrid_il_rl} we provide the details, with DAgger on \hi level and $Q$-learning on $\lo$ level. The scheme can be adapted to other interactive IL and RL algorithms.

The learning agent proceeds by \textit{rolling in} with its meta-controller (line 7). For each selected subgoal $g$, the subpolicy $\pi_g$ selects and executes primitive actions via the $\epsilon$-greedy rule (lines 9--10), until some termination condition is met. The agent receives some pseudo-reward, also known as intrinsic reward~\citep{kulkarni2016hierarchical} (line 10).
Upon termination of the subgoal, agent's meta-controller $\mu$ chooses another subgoal and the process continues until the end of the episode, where the involvement of the expert begins.
As in hg-DAgger, the expert inspects the overall execution of the learner (line 17), and if it is not successful,
the expert provides \hi-level labels, which are accumulated for training the meta-controller.

Hierarchical guidance impacts how the \lo-level learners accumulate experience. As long as the meta-controller's subgoal $g$ agrees with the expert's, the agent's experience of executing subgoal $g$ is added to the experience replay buffer $\Dg$. If the meta-controller selects a ``bad'' subgoal, the accumulation of experience in the current episode is terminated. This ensures that experience buffers contain only the data from the relevant part of the state space.

Algorithm~\ref{alg:hybrid_il_rl} assumes access to a real-valued function $\pseudo(s;g)$, providing the pseudo-reward in state $s$ when executing $g$, and
a predicate $\terminal(s;g)$, indicating the termination (not necessarily successful) of subgoal~$g$. This setup is similar
to prior work on hierarchical RL~\cite{kulkarni2016hierarchical}. One natural definition of pseudo-rewards, based on an additional predicate $\success(s;g)$ indicating a successful completion of subgoal $g$, is as follows:
%The $\pseudo$ provides intrinsic reward for the agent based on 3 scenarios (line 13):
\[
%\tilde{r}\gets
\begin{cases}
1&\text{if } \success(s;g)
\\
-1&\text{if } \neg\success(s;g) \text{ and } \terminal(s;g)
\\
-\timepen&\text{otherwise,}
\end{cases}
\]
where $\timepen>0$ is a small penalty to encourage short trajectories.
The predicates \success and \terminal are provided by an expert or learnt from supervised or reinforcement feedback.
In our experiments, we explicitly provide these predicates to both hg-DAgger/Q as well as the hierarchical RL, giving them
advantage over hg-DAgger, which needs to learn when to terminate subpolicies.

%In principle, one could learn when to terminate purely by reinforcement learning, by treating termination as part of the augmented action space.  However
%the sample complexity for the reinforcement learners at the \lo level would only increase---thus, we are giving the IL--RL hybrid (and hierarchical RL) an advantage over hierarchical DAgger in our comparisons. In practice, the termination condition might be learned from a separate supervised (passive or active) feedback. We defer learning of the termination function for future work.

%In principle, one could instead attempt to learn when to terminate, treating termination as part of the augmented action space, but
%the sample complexity for the reinforcement learners at the \lo level tends to be significantly higher compared to imitation learning, slowing down the progress and requiring significantly more access to the expert.

%Algorithm \ref{alg:hybrid_il_rl} applies to scenarios where expert feedback is given only at the higher level, and $\lo$-level policies are learned from conventional reinforcement learning techniques. In this case, the $\texttt{pseudo-reward}$ function supplies the training signal for when the $\lo$-level policy $\pi_g$ successfully completes a subtask $g$.

%\mdcomment{Keeping the old algorithms / descriptions for reference for now}
{%
}
%%% Local Variables:
%%% mode: latex
%%% TeX-master: "icml_paper.tex"
%%% End:

\section{Experiments} \label{sec:experiments} \makeatletter
\newcommand{\customlabel}[3]{%
%%% without hyperref: only 2 arguments instead of 5
   \protected@write \@auxout {}{\string \newlabel {#1}{{\ref{#2}~(#3)}{\thepage}{Subfigure #2 (#3)\relax}{}{}} }%
   \protected@write \@auxout {}{\string \newlabel {sub@#1}{{#3}{\thepage}{Subfigure #2 (#3)\relax}{}{}} }%
}
\makeatother

\begin{figure*}
	\centering     %%% not \center
\hfill
    \customlabel{fig:env_screenshot}{fig:maze}{left}%
		\includegraphics[height=1.70in]{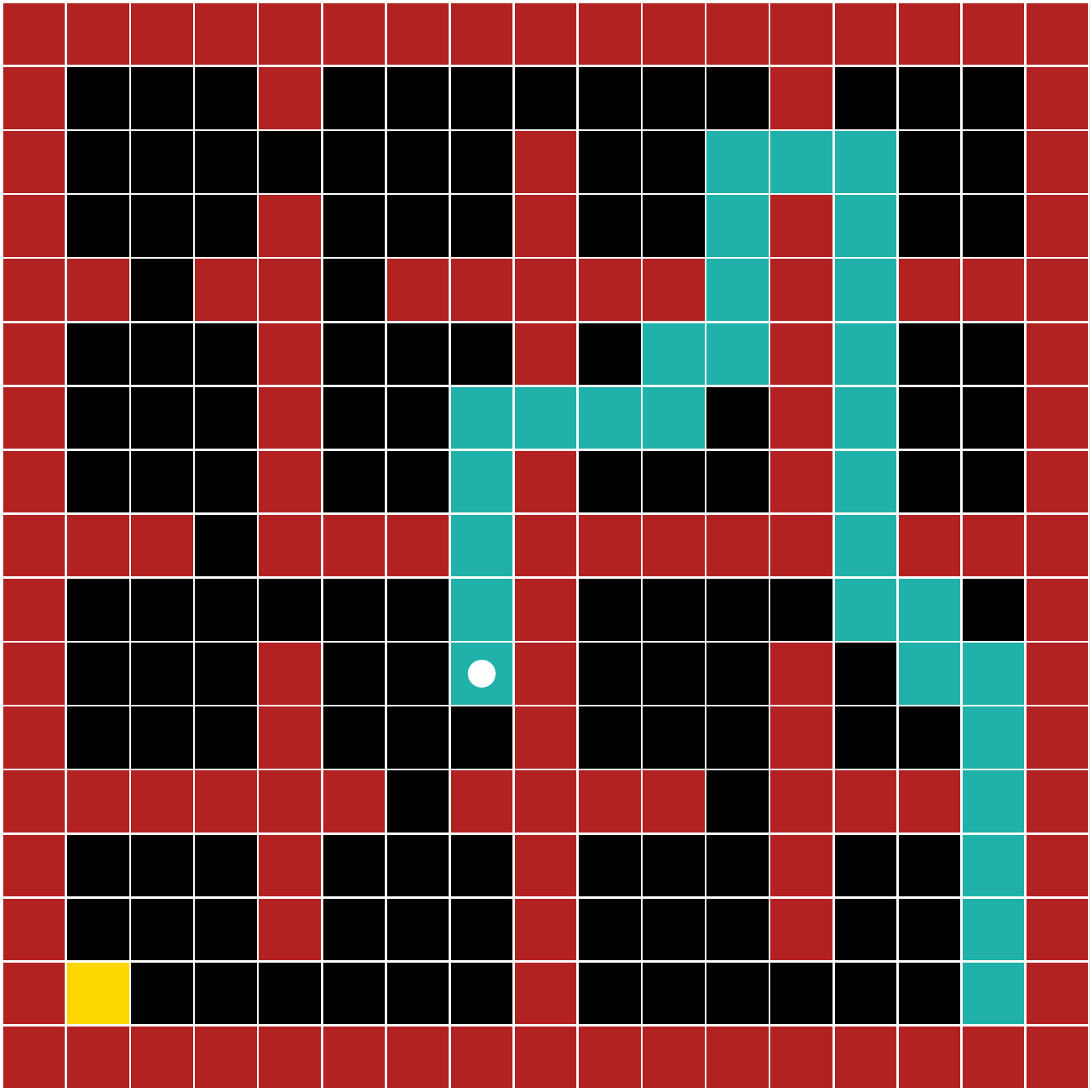}
\hfill%
    \customlabel{fig:labels_by_type}{fig:maze}{middle}%
		\includegraphics[height=1.80in]{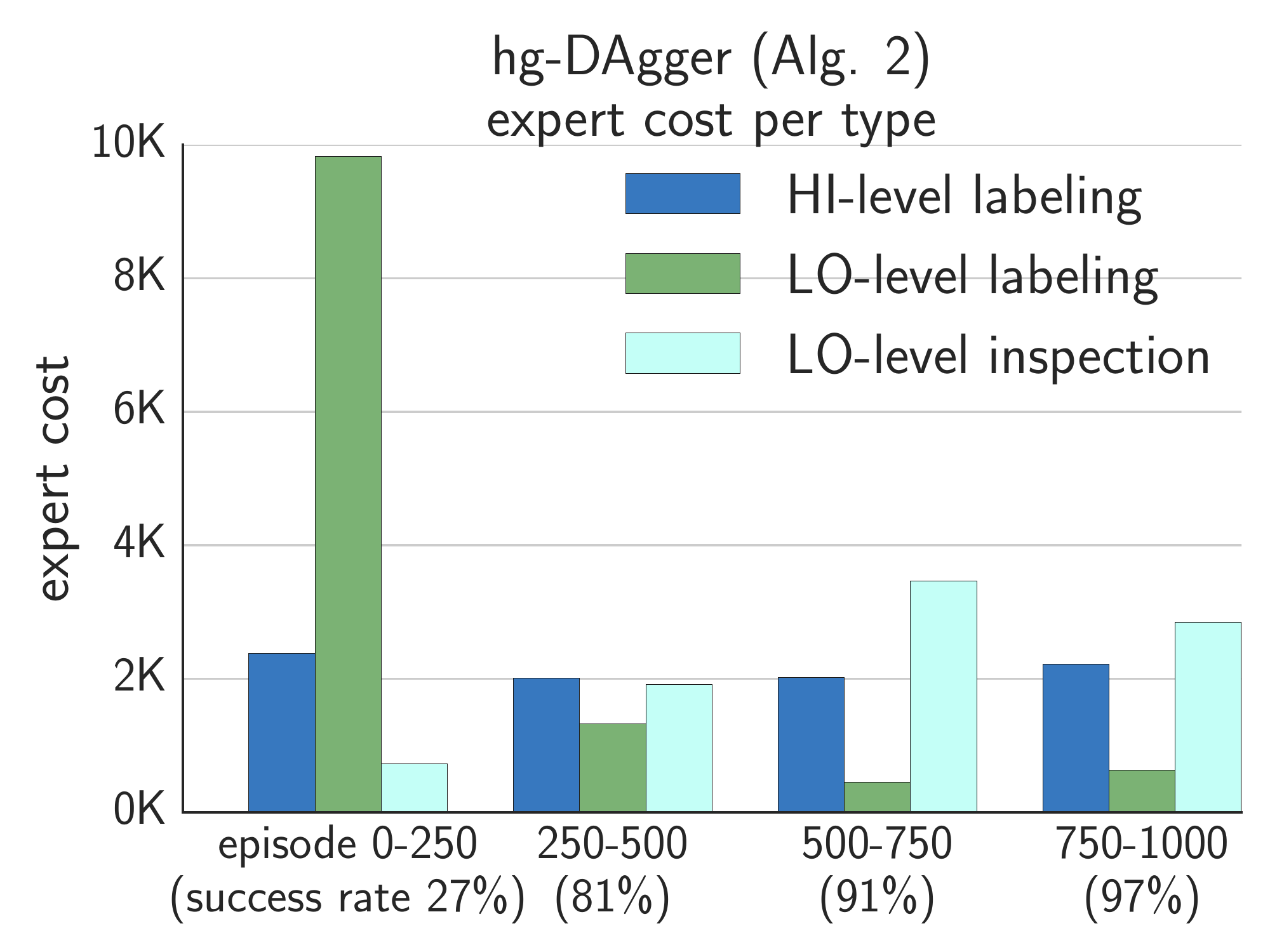}
\hfill%
    \customlabel{fig:hybrid_rl_il}{fig:maze}{right}%
		\includegraphics[height=1.80in]{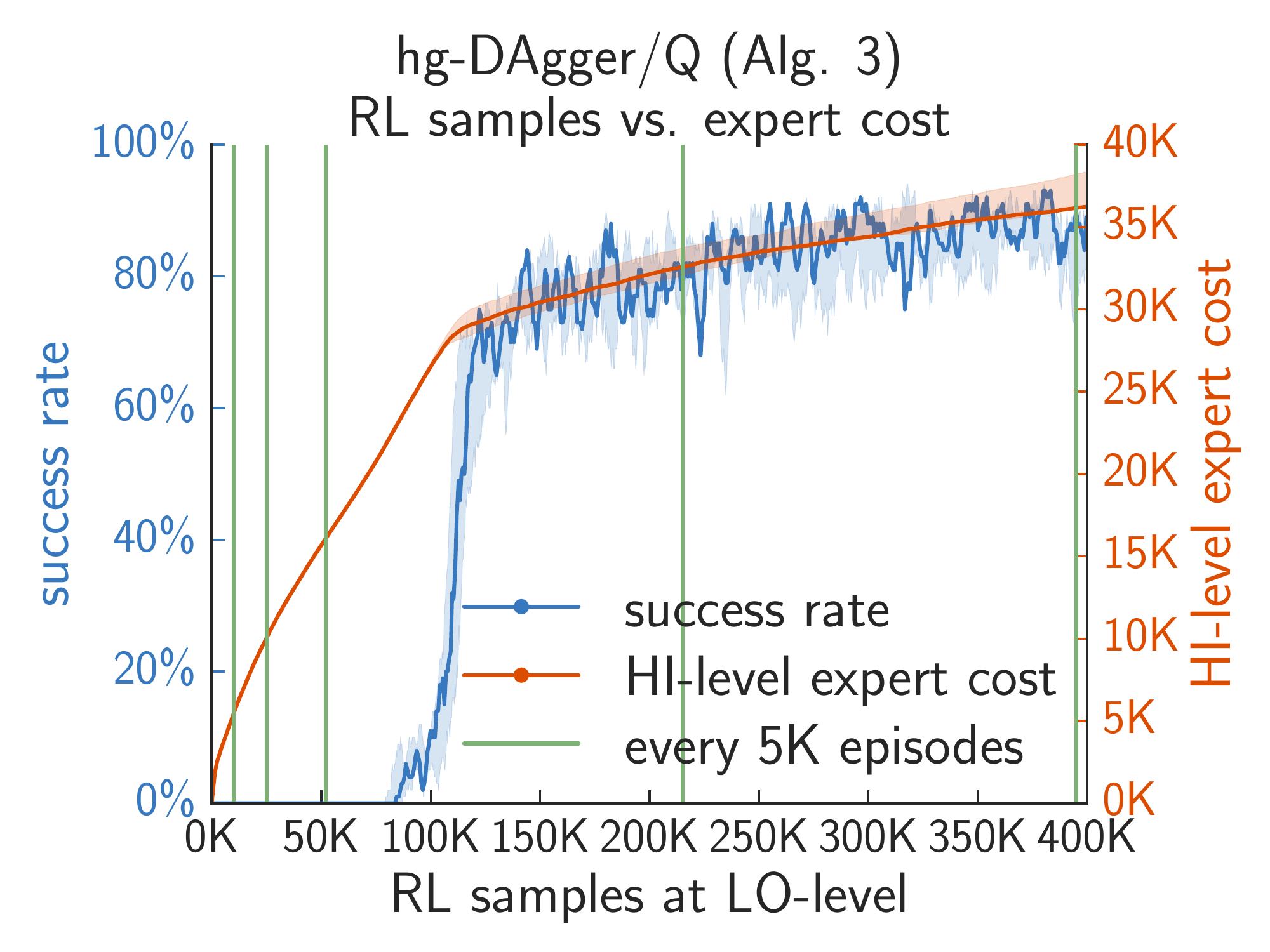}
        \vspace{-0.1in}
	\caption{\emph{Maze navigation}.
             \emph{(Left)} One sampled environment instance; the agent needs to navigate from bottom right to bottom left.
             \emph{(Middle)} Expert cost over time for hg-DAgger; the cost of $\lbl$ operations equals the length
             of labeled trajectory, the cost of $\chk$ operations is 1.
             %counting all of $\lbllo, \lblhi$ and $\lblfu$ on the same unit cost scale.
             \emph{(Right)} Success rate of hg-DAgger/Q and the \hi-level label cost
             %(i.e., the number of subgoals generated by expert)
             as a function of the number of \lo-level RL samples.}
\label{fig:maze}
\end{figure*}

\begin{figure*}
	\centering     %%% not \center
%	\captionsetup{justification=centering}
%	\subfigure[]{\label{fig:episode_success_indicator}\includegraphics[width=55mm]{episode_success_indicator.png}}
%	\subfigure[]{\label{fig:label_complexity}\includegraphics[width=55mm]{label_complexity.png}}
%	\subfigure[]{\label{fig:L2_comparison}\includegraphics[width=55mm]{expert_L2_comparison.png}}
%	\vskip -0.2in
    \customlabel{fig:episode_success_indicator}{fig:cmp:flat}{left}%
        \includegraphics[width=0.33\textwidth]{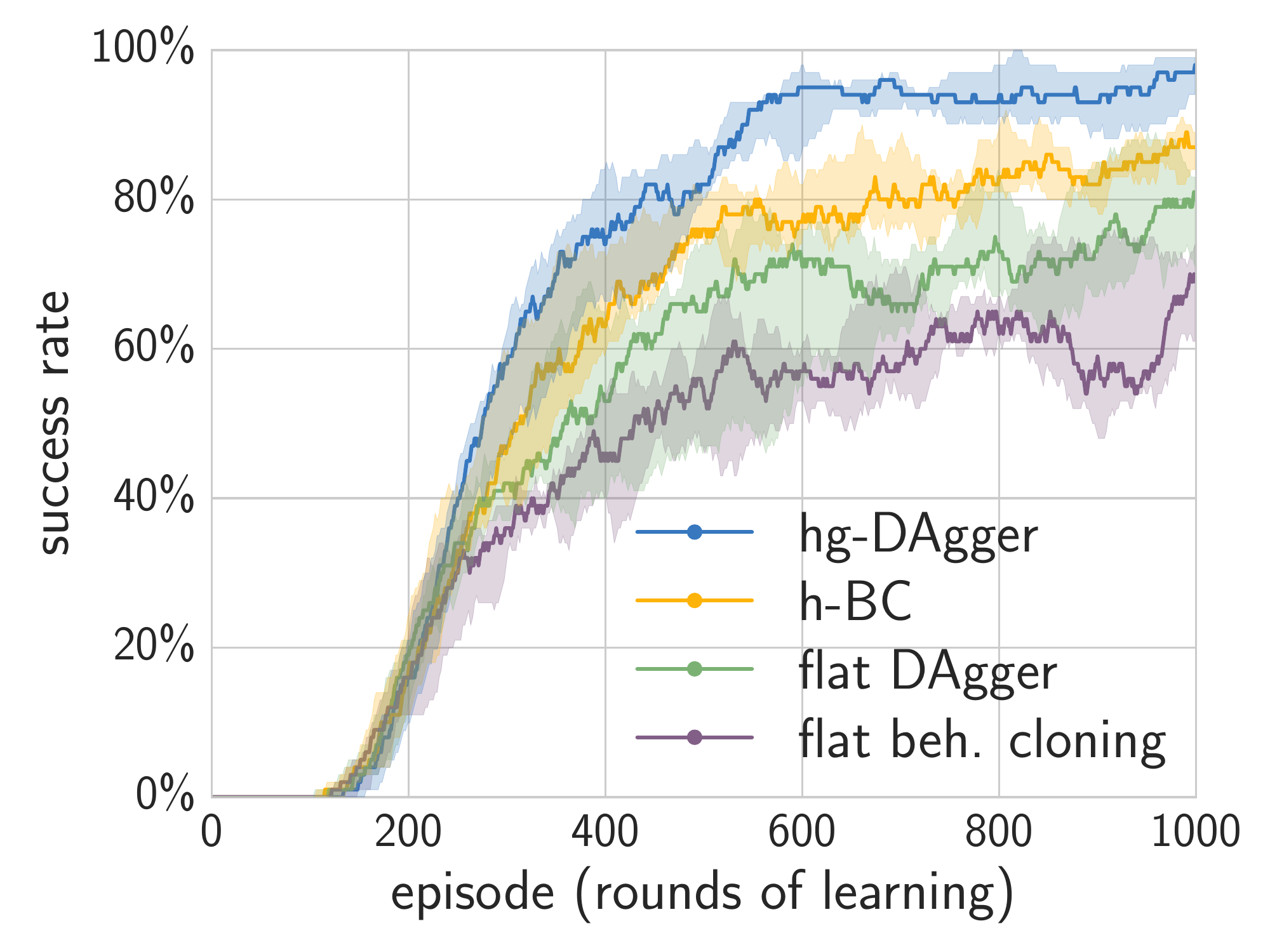}%
\hfill%
    \customlabel{fig:label_complexity}{fig:cmp:flat}{middle}%
        \includegraphics[width=0.33\textwidth]{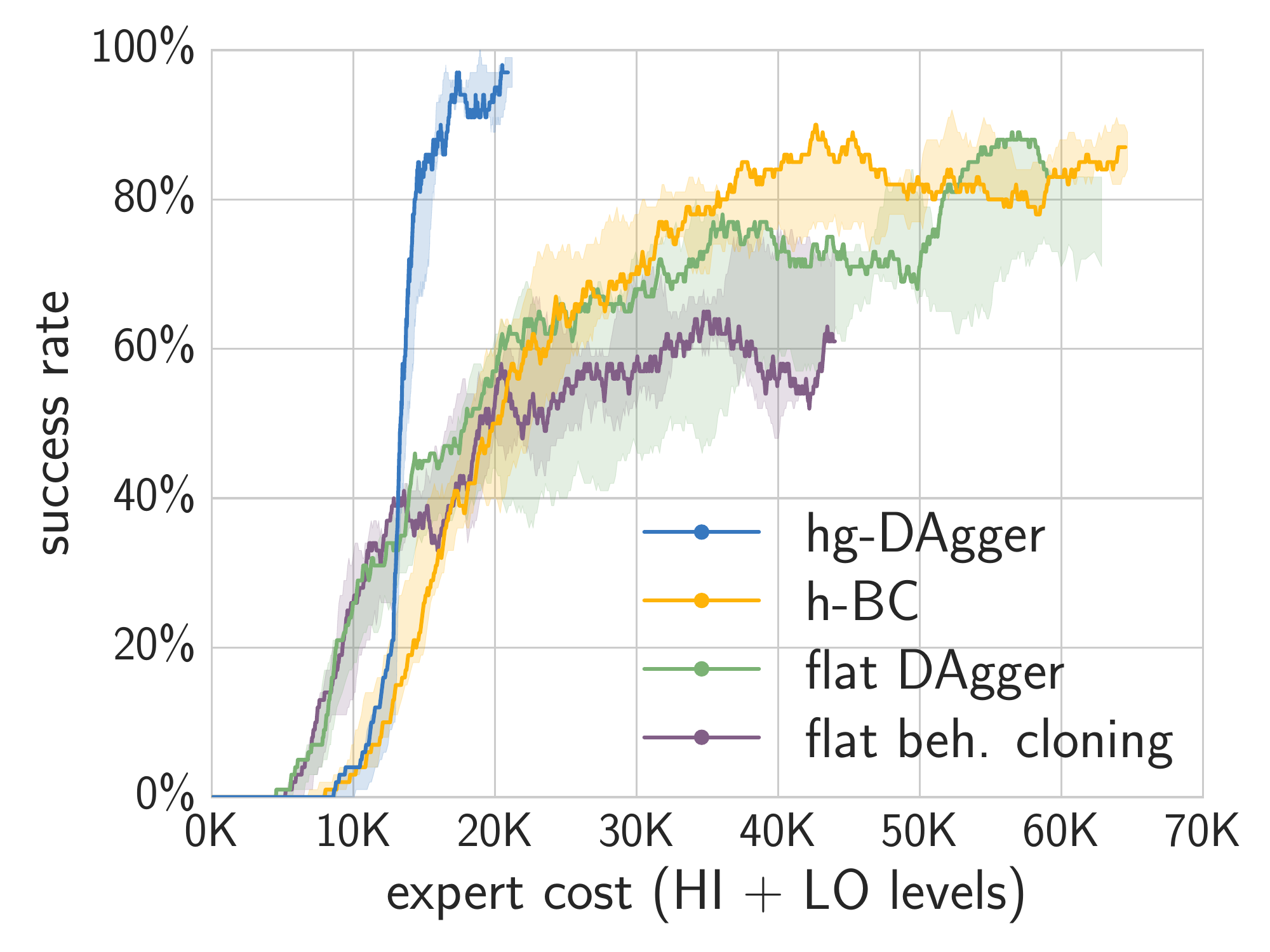}%
\hfill%
    \customlabel{fig:L2_comparison}{fig:cmp:flat}{right}%
        \includegraphics[width=0.33\textwidth]{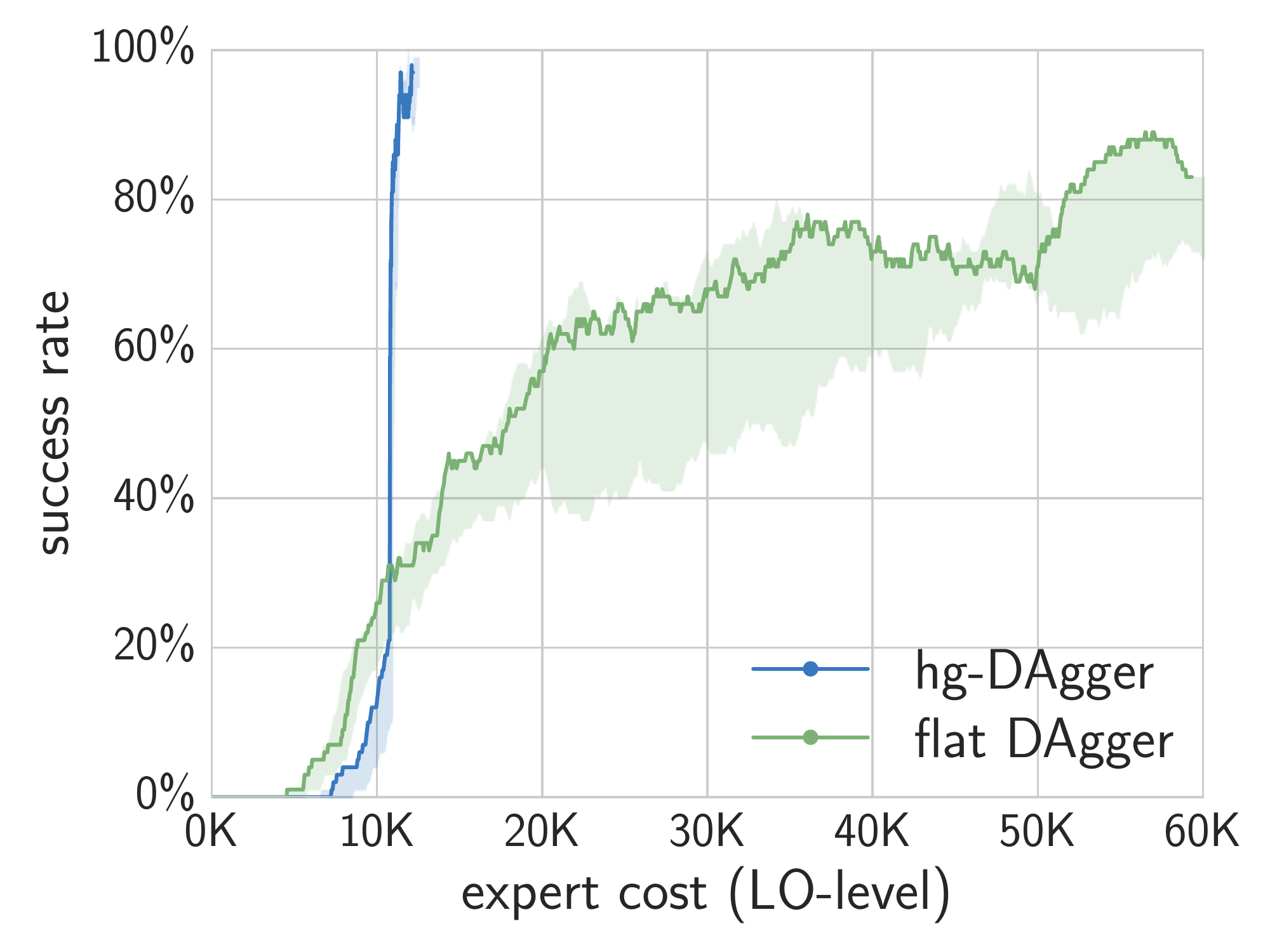}%
        \vspace{-0.1in}
	\caption{\emph{Maze navigation: hierarchical versus flat imitation learning}. Each episode is followed by a round of training and a round of testing. The success rate is measured over previous 100 test episodes; the expert cost is as in Figure~\ref{fig:maze}.
%labeling effort is measured by the number of expert labels, i.e., the number of subgoals or primitive actions generated by the expert.
%Thus, the cost of each $\lbl$ operation is equal to the length of the labeled trajectory.
             \emph{(Left)} Success rate per episode.
             \emph{(Middle)} Success rate versus the expert cost.
             %counting all of $\lbllo, \lblhi$ and $\lblfu$ on the same unit cost scale.
             \emph{(Right)} Success rate versus the \lo-level expert cost.}
%---obtained by $\lbllo$ for hierarchical DAgger and $\lblfu$ for flat DAgger.}
    \label{fig:cmp:flat}
\end{figure*}

%\begin{verbatim}
%Maze
% - (only one scale of the environment)
% - IL/IL, flat-IL, IL/RL:
%    - Success rate versus episode
%    - Success rate versus labeling cost (assuming a unique cost per action-level label)
% - Flat-RL and RL/RL
%    - Try to run with a handicap: run expert for T0 steps and then start RL
%    - T0=?
%    - [these are lower priority experiments; if experiments not ready, we can just say that the long horizon on high level is too prohibitive for both of these]
% - For IL/IL: Show the #calls to different types of queries over time

%Montezuma
% - Flat-RL doesn't work
% - IL/RL versus RL/RL:
%    - Success rate versus episode; show #expert labels on the same axis
%    - Make a point that this environment should be easy for the high-level RL (because of low effective branching & relatively short time horizon)
%    - For IL/RL hybrid learning of different options over time
%\end{verbatim}
We evaluate the performance of our algorithms on two separate domains: (i) a simple but challenging maze navigation domain and (ii) the Atari game Montezuma's Revenge.
%Both domains require sufficiently long planning horizons that flat reinforcement learning techniques such as Deep $Q$-Learning \cite{mnih2015human} are intractable.

\subsection{Maze Navigation Domain}
\label{overview}
%The setup is one of random encounter with a large number of deterministic MDPs.

\textbf{Task Overview.} Figure~\ref{fig:env_screenshot} displays a snapshot of the maze navigation domain.
%The setup is one of random encounter with a large number of deterministic MDPs.
%
In each episode, the agent encounters a new instance of the maze from a large collection of different layouts. Each maze consists of 16 rooms arranged in a 4-by-4 grid, but the openings between the rooms vary from instance to instance as does the initial position of the agent and the target. The agent (white dot) needs to navigate from one corner of the maze to the target marked in yellow. Red cells are obstacles (lava), which the agent needs to avoid for survival. The contextual information the agent receives is the pixel representation of a bird's-eye view of the environment, including the partial trail (marked in green) indicating the visited locations.

Due to a large number of random environment instances, this domain is not solvable with tabular algorithms. Note that rooms are not always connected, and the locations of the hallways are not always in the middle of the wall. Primitive actions include going one step \emph{up}, \emph{down}, \emph{left} or \emph{right}. In addition, each instance of the environment is designed to ensure that there is a path from initial location to target, and the shortest path takes at least 45 steps ($\Hfu =100$). The agent is penalized with reward $-1$ if it runs into lava, which also terminates the episode. The agent only receives positive reward upon stepping on the yellow block. %As such, solving this environment using pure exploration-based reinforcement learning approaches such as DQN will be highly challenging due to long planning horizon and sparse rewards.

A hierarchical decomposition of the environment corresponds to four possible subgoals of going to the room immediately to the \emph{north}, \emph{south}, \emph{west}, \emph{east}, and the fifth possible subgoal \emph{go\hspace{-1pt}\_\hspace{1pt}to\hspace{-1pt}\_\hspace{1pt}target} (valid only in the room containing the target). %\aacomment{Not sure if I follow this. So if going north takes you away from the goal, then it is not applicable as a subgoal in the current state?}
In this setup, $\Hlo\approx\text{5 steps}$, and $\Hhi\approx\text{10--12 steps}$. The episode terminates after 100 primitive steps if the agent is unsuccessful. The subpolicies and meta-controller use similar neural network architectures and only differ in the number of action outputs. (Details of network architecture are provided in Appendix~\ref{sec:app_experiment}.)

%\subsection{Results on Maze Domain}
%\begin{figure}[h]
%	\vskip 0.0in
%	\begin{center}
%		\centerline{\includegraphics[width=0.6\columnwidth]{hierarchical_dagger_labels_breakdown.png}}
%		\caption{\textit{Types of feedback over time for Hierarchical DAgger}}
%		\label{fig:labels_by_type}
%		%			\vskip -0.2in
%	\end{center}
%	%	\vskip -0.2in
%\end{figure}

\textbf{Hierarchically Guided IL.}
We first compare our hierarchical IL algorithms with their flat versions. The algorithm performance is measured by success rate, defined as the average rate of successful task completion over the previous 100 test episodes, on random environment instances not used for training. The cost of each $\lbl$ operation equals the length of the labeled trajectory, and the cost of each $\chk$ operation equals 1.

Both h-BC and hg-DAgger outperform flat imitation learners (\Fig{cmp:flat}, \subref{fig:episode_success_indicator}). hg-DAgger, in particular, achieves consistently the highest success rate, approaching 100\% in fewer than 1000 episodes. \Fig{episode_success_indicator} displays the median as well as the range from minimum to maximum success rate over 5 random executions of the algorithms.

Expert cost varies significantly between the two hierarchical algorithms. Figure \ref{fig:label_complexity} displays the same success rate, but as a function of the expert cost.
%, where we assign the same unit cost to both \lo and \hi-level operations for simplicity of comparison.}\sout{total number of expert labels.}
%
%Here to simplify the comparison, we treat each \hi-level label cost to be the same as each \lo-level label cost.
%
hg-DAgger achieves significant savings in expert cost compared to other imitation learning algorithms thanks to a more efficient use of the \lo-level expert through hierarchical guidance. Figure \ref{fig:labels_by_type} shows that hg-DAgger requires most of its $\lo$-level labels early in the training and requests primarily $\hi$-level labels after the subgoals have been mastered. As a result, hg-DAgger requires only a fraction of \lo-level labels compared to flat DAgger (Figure~\ref{fig:cmp:flat}, \subref{fig:L2_comparison}).
%When the cost of providing feedback at the $\lo$ level is high, namely $\CLhi \ll \CLlo \approx \CLfu$, Figure \ref{fig:L2_comparison} indicates that Hierarchical DAgger should achieve saving of labeling cost through significant reduction in the required labeling at $\lo$ level. \aacomment{$\CLfu$ should be around $H_{\hi} \CLlo$, no?}

\textbf{Hierarchically Guided IL\,/\,RL.}
We evaluate hg-DAgger/Q with deep double $Q$-learning (DDQN, \citealp{van2016deep}) and prioritized experience replay \cite{schaul2015prioritized} as the underlying RL procedure.
Each subpolicy learner receives a pseudo-reward of 1 for each successful execution, corresponding to stepping through the correct door (e.g., door to the north if the subgoal is \emph{north}) and negative reward for stepping into lava or through other doors.

\Fig{hybrid_rl_il} shows the learning progression of hg-DAgger/Q, implying two main observations.
First, the number of \hi-level labels rapidly increases initially and then flattens out after the learner becomes more successful, thanks to the availability of $\chkfu$ operation. As the hybrid algorithm makes progress and the learning agent passes the $\chkfu$ operation increasingly often, the algorithm starts saving significantly on expert feedback.
Second, the number of \hi-level labels is higher than for both hg-DAgger and h-BC. $\chkfu$ returns \textit{Fail} often, especially during the early parts of training. This is primarily due to the slower learning speed of $Q$-learning at the \lo level, requiring more expert feedback at the \hi level.
This means that the hybrid algorithm is suited for settings where \lo-level expert labels are either not available or more expensive than the \hi-level labels. This is exactly the
setting we analyze in the next section.

In Appendix~\ref{app:maze}, we compare hg-DAgger/Q with hierarchical RL (h-DQN, \citealp{kulkarni2016hierarchical}), concluding that h-DQN, even with significantly more \lo-level samples, fails to reach success rate comparable to hg-DAgger/Q. Flat $Q$-learning also fails in this setting, due to a long planning horizon and sparse rewards~\cite{mnih2015human}.

%
%This combination of two techniques has demonstrated improvement over standard DQN baseline.
%enhance the learning procedure for h-DQN with double learning and prioritized experience replay.

%and in the next section focus on the comparison of hybrid IL-RL and hierarchical RL in the Montezuma's Revenge domain.

\subsection{Hierarchically Guided IL\,/\,RL vs Hierarchical RL: Comparison on Montezuma's Revenge}

\begin{figure*}
	\centering     %%% not \center
%	\captionsetup{justification=centering}
%	\subfigure[]{\label{fig:montezuma_screen}\includegraphics[width=55mm]{atari_subgoal_sequence.png}}
%	\subfigure[]{\label{fig:atari_progression}\includegraphics[width=55mm]{atari_learning_progression.png}}
%	\subfigure[]{\label{fig:atari_hybrid_vs_hdqn}\includegraphics[width=55mm]{atari_hybrid_vs_hdqn.png}}
%	\vskip -0.2in
    \customlabel{fig:montezuma_screen}{fig:montezuma}{left}%
        \includegraphics[width=0.33\textwidth]{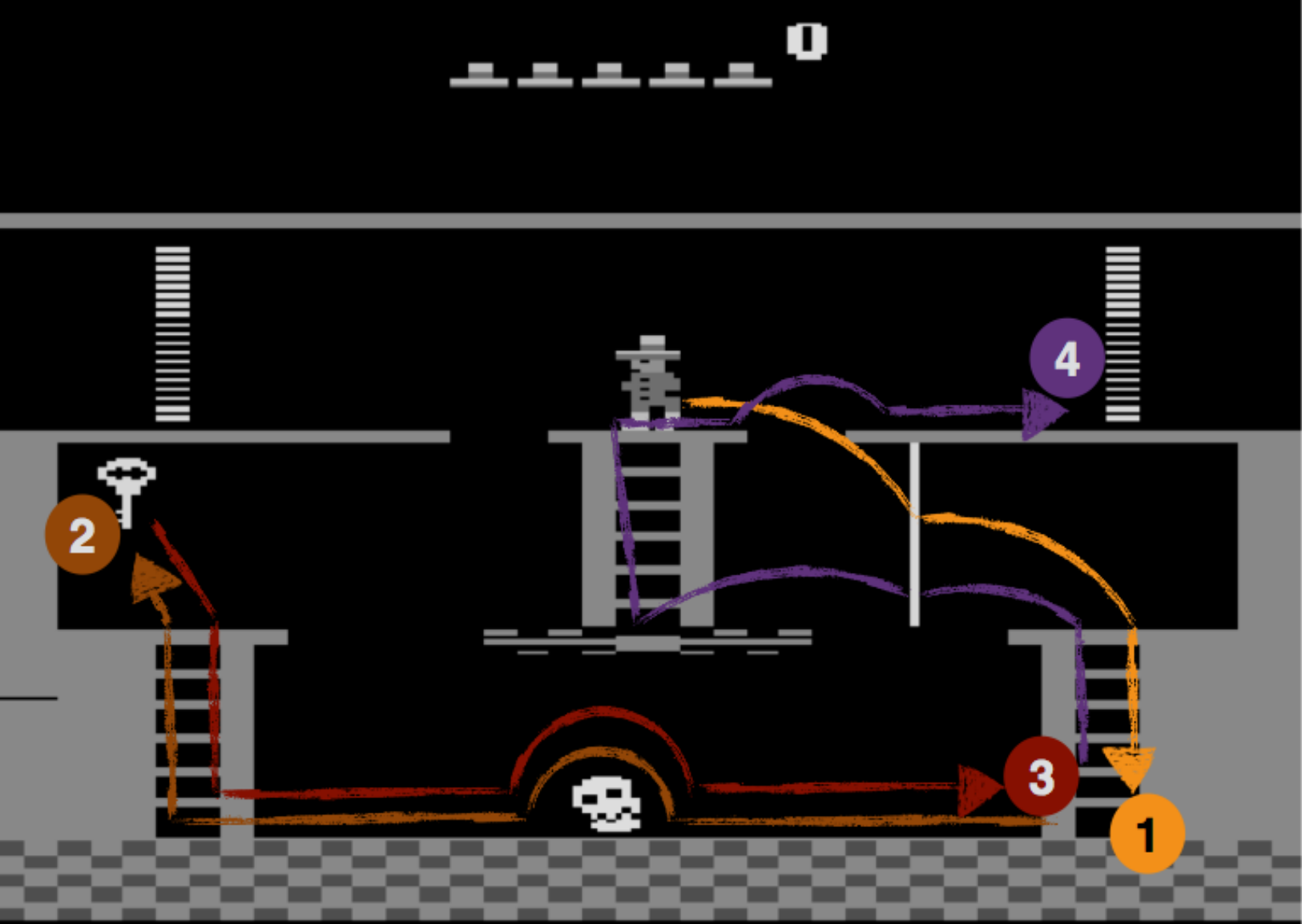}%
\hfill%
    \customlabel{fig:atari_progression}{fig:montezuma}{middle}%
        \includegraphics[width=0.33\textwidth]{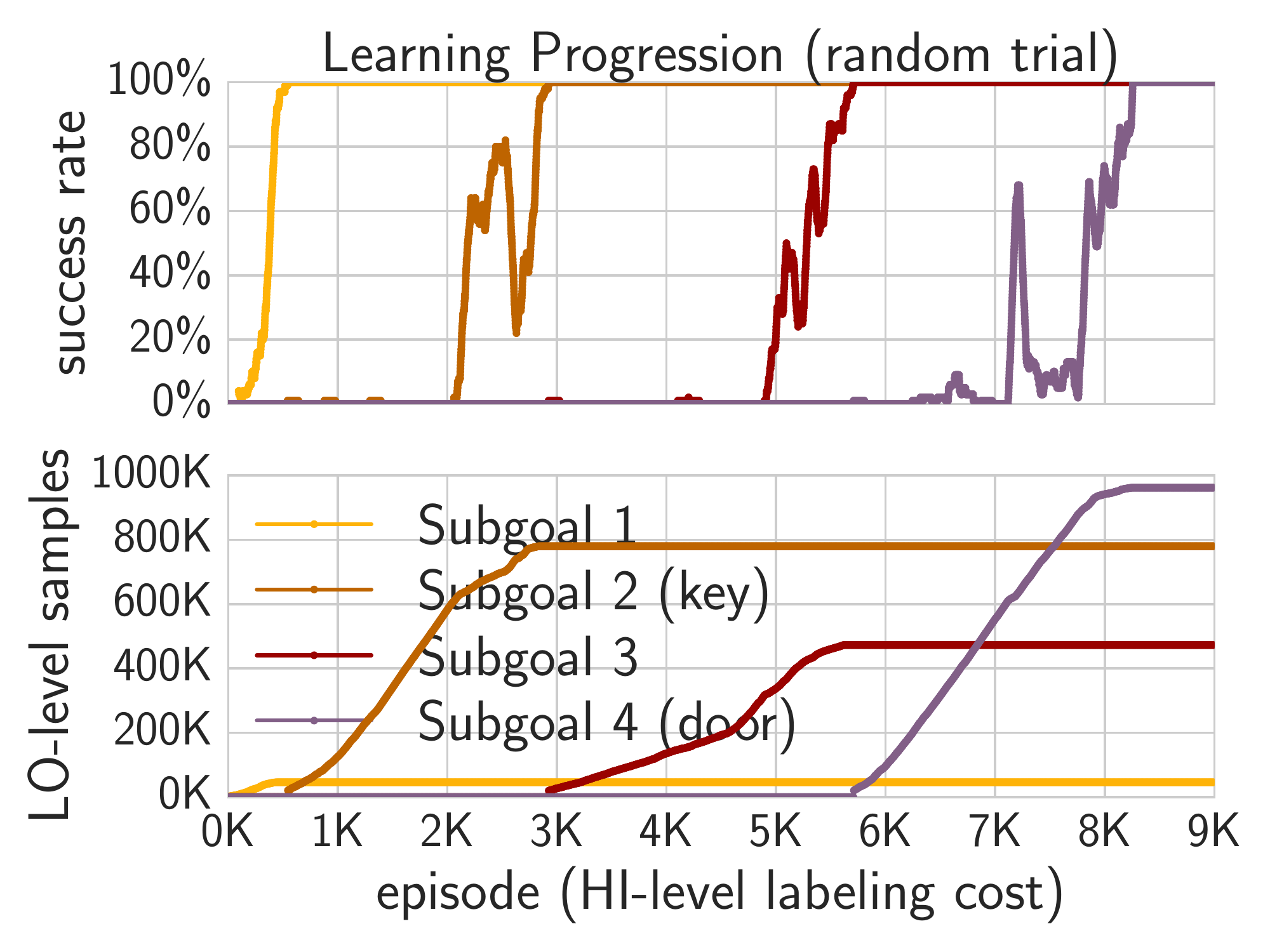}%
\hfill%
    \customlabel{fig:atari_hybrid_vs_hdqn}{fig:montezuma}{right}%
        \includegraphics[width=0.33\textwidth]{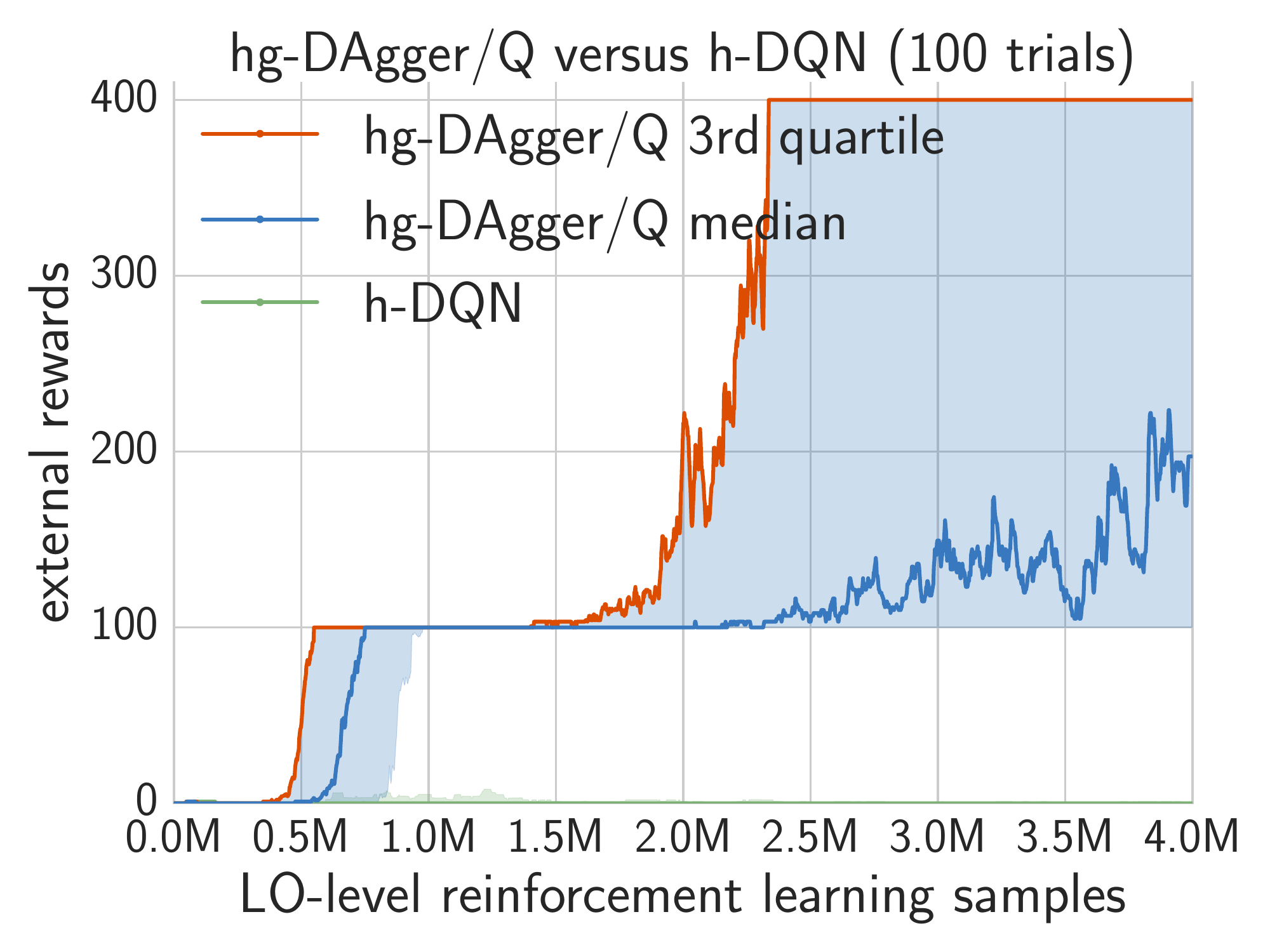}%
        \vspace{-0.1in}
	\caption{\emph{Montezuma's revenge: hg-DAgger/Q versus h-DQN.}
             \emph{(Left)} Screenshot of Montezuma's Revenge in black-and-white with color-coded subgoals.
             \emph{(Middle)} Learning progression of hg-DAgger/Q in solving the first room of Montezuma's Revenge for a typical successful trial. Subgoal colors match the left pane; success rate is the fraction of times the \lo-level RL learner achieves its subgoal over the previous 100 attempts.
             \emph{(Right)} Learning performance of hg-DAgger/Q versus h-DQN (median and inter-quartile range).}
\label{fig:montezuma}
\end{figure*}

\textbf{Task Overview.}
Montezuma's Revenge is \edit{among the most difficult Atari games}\sout{among the Atari games that are the most difficult} for existing deep RL algorithms, and is a natural candidate for hierarchical approach due to the sequential order of subtasks. Figure \ref{fig:montezuma_screen} displays the environment and an annotated sequence of subgoals. The four designated subgoals are: go to bottom of the right stair, get the key, reverse path to go back to the right stair, then go to open the door (while avoiding obstacles throughout).
%, see color coded \Fig{montezuma}, \subref{fig:montezuma_screen}).

The agent is given a pseudo-reward of 1 for each subgoal completion and -1 upon loss of life. We enforce that the agent can only have a single life per episode, preventing the agent from taking a shortcut after collecting the key (by taking its own life and re-initializing with a new life at the starting position, effectively collapsing the task horizon). Note that for this setting, the actual game environment is equipped with two positive external rewards corresponding to picking up the key (subgoal 2, reward of 100) and using the key to open the door (subgoal 4, reward of 300). Optimal execution of this sequence of subgoals requires more than 200 primitive actions. Unsurprisingly, flat RL algorithms often achieve a score of $0$ on this domain \cite{mnih2015human, mnih2016asynchronous, wang2016dueling}.

%\begin{figure}[h]
%	\vskip 0.0in
%	\begin{center}
%		\centerline{\includegraphics[width=0.6\columnwidth]{atari_subgoal_sequence.png}}
%		\caption{Screenshot of Montezuma's Revenge in black \& white. Subgoals are color coded corresponding to figure \ref{fig:atari_progression}. }
%		\label{fig:montezuma_screen}
%	\end{center}
%	\vskip -0.2in
%\end{figure}
%\aacomment{How exactly are pseudo-rewards allocated? I am confused by the rewards being 100, 400 etc. I thought that the pseudo-rewards were only 1 in our algorithm, and we should describe if they are not here and possibly make it a parameter in the algorithm.} \aacomment{Are the other subgaols easy to describe as well?}

\textbf{hg-DAgger/Q versus h-DQN.}
Similar to the maze domain, we use DDQN with prioritized experience replay at the $\lo$ level of hg-DAgger/Q. We compare its performance with
h-DQN using the same neural network architecture as~\citet{kulkarni2016hierarchical}.
Figure \ref{fig:atari_progression} shows the learning progression of our hybrid algorithm. The \hi-level horizon $H_\hi = 4$, so meta-controller is learnt from fairly few samples. Each episode roughly corresponds to one $\lblhi$ query. Subpolicies are learnt in the order of subgoal execution as prescribed by the expert.

We introduce a simple modification to $Q$-learning on the \lo level to speed up learning: the accumulation of experience replay buffer does not begin until the first time the agent encounters positive pseudo-reward.
%
%This modification is well-suited for the considered interaction mode where expert is giving advice at the $\hi$ level (one can imagine the expert simply indicates when the $\lo$-level learning should begin). \sout{This mechanism explains, for example, the temporal gap between mastering subgoal 3 and commencement of learning subgoal 4 in \Fig{atari_progression}.}
%
During this period, in effect, only the meta-controller is being trained. This modification ensures the reinforcement learner encounters at least some positive pseudo-rewards, which boosts learning in the long horizon settings and should naturally work with any off-policy learning scheme (DQN, DDQN, Dueling-DQN). For a fair comparison, we introduce the same modification to the h-DQN learner (otherwise, h-DQN failed to achieve any reward).

To mitigate the instability of DQN (see, for example, learning progression of subgoal 2 and 4 in Figure \ref{fig:montezuma}, middle),
we introduce one additional modification.
We terminate training of subpolicies when the success rate exceeds 90\%, at which point the subgoal is considered learned. Subgoal success rate is defined as the percentage of successful subgoal completions over the previous 100 attempts.
%The termination of subgoal training is a practical way to cope with the instability of DQN
%For Montezuma's Revenge domain, setting termination threshold is also practical as the \lo and \hi level learners have significantly different learning speed.
%% MD: I excluded the sentence, because I don't follow. We might re-include after rephrasing

%Similar to \cite{kulkarni2016hierarchical}, we evaluate the sample complexity of solving the entire first room of Montezuma's Revenge. We compare our hybrid algorithm with h-DQN enhanced by our modifications. Note that h-DQN fails to achieve any reward without these enhancements.
Figure \ref{fig:atari_hybrid_vs_hdqn} shows the median and the inter-quartile range over 100 runs of hg-DAgger/Q and hg-DQN.\footnote{%
In Appendix~\ref{sec:app_experiment}, we present additional plots, including 10 best runs of each algorithm, subgoal completion rate over 100 trials, and versions of Figure \ref{fig:atari_progression} for additional random instances.}
The \lo-level sample sizes are not directly comparable with the middle panel, which displays the learning progression for a random successful run, rather than an aggregate over multiple runs.
%We include additional results, with some failing examples of our own reinforcement learning procedure, in the appendix.
%
In all of our experiments, the performance of the imitation learning component is stable across many different trials, whereas the performance of the reinforcement learning component varies substantially. Subgoal 4 (door) is the most difficult to learn due to its long horizon whereas subgoals 1--3 are mastered very quickly, especially compared to h-DQN. Our algorithm benefits from hierarchical guidance and accumulates experience for each subgoal only within the relevant part of the state space, where the subgoal is part of an optimal trajectory. In contrast, h-DQN may pick bad subgoals and the resulting \lo-level samples then ``corrupt'' the subgoal experience replay buffers and substantially slow down convergence.\footnote{%
In fact, we further reduced the number of subgoals of h-DQN to only two initial subgoals, but the agent still largely failed to learn even the second subgoal (see the appendix for details).} %This is in line with the observations of~\citet{roderick2017abstract}.}
%
%Intuitively, the sequential nature of subgoals in this setting makes it difficult to learn multiple subpolicies simultaneously without mutually corrupting the experience replay buffers.
%The use of imitation learning at the $\hi$ level naturally streamlines multiple learning tasks.

\edit{The number of \hi-level labels in Figure \ref{fig:atari_progression} can be further reduced by using a more efficient RL procedure than DDQN at the \lo level. In the specific example of Montezuma's Revenge, the actual human effort is in fact much smaller, since the human expert needs to provide a sequence of subgoals only once (together with simple subgoal detectors), and then \hi-level labeling can be done automatically. The human expert only needs to understand the high level semantics, and does not need to be able to play the game.}
%%% Local Variables:
%%% mode: latex
%%% TeX-master: "icml_paper.tex"
%%% End:

\vspace{-0.1in}
\section{Conclusion} \label{sec:discussion} \edit{We have presented \emph{hierarchical guidance} framework and shown how it can be used to speed up learning and reduce the cost of expert feedback in
hierarchical imitation learning and hybrid imitation--reinforcement learning.}

%Our approach is flexible and can be instantiated to incorporate a mixture of imitation and reinforcement feedback at different levels of the hierarchy.  Compared to flat imitation learning, our approach enjoys significantly improved sample complexity, both theoretically and empirically. Compared to hierarchical reinforcement learning, our approach achieves significantly faster convergence in practice.

Our approach can be extended in several ways. For instance, one can consider weaker feedback such as preference or gradient-style feedback \cite{furnkranz2012preference,loftin2016learning,christiano2017deep},
or a weaker form of imitation feedback, only saying whether the agent action is correct or incorrect, corresponding to bandit variant of imitation learning~\cite{ross2011reduction}.

Our hybrid IL\,/\,RL approach relied on the availability of a subgoal termination predicate indicating when the subgoal is achieved.  While in many settings such a termination predicate is relatively easy to specify, in other settings this predicate needs to be learned. We leave the question of learning the termination predicate, while learning to act from reinforcement feedback, open for future research.

%Finally, one can also study further variations of our framework.  For instance, one could employ reinforcement learning at the top level and imitation learning at the bottom level.  One reason we did not study this variant was the lack of a good motivating setting.  Another direction is to extend beyond two levels or study the multi-task or transfer setting.

%Other forms of supervision/weaker teachers: yes / no, better/worse, policy gradient style feedback (M. Littman)

%Discuss the existence of RL/IL hybrid--open for future work

%Subgoal detectors / termination predicate: can be learned. Likely requires much less data than RL, because it's just a classification task

%More than two levels

%%% Local Variables:
%%% mode: latex
%%% TeX-master: "icml_paper.tex"
%%% End:

\section*{Acknowledgments}

The majority of this work was done while HML was an intern at Microsoft Research.  HML is also supported in part by an Amazon AI Fellowship.

%\input{sec_intro}
%\input{sec_hierarchical_il}
%\input{sec_analysis}
%\input{sec_hybrid}
%\input{sec_maze}
%\input{sec_atari}
%\input{sec_related}

%\begin{small}
\bibliography{icml_paper}
\bibliographystyle{icml2018}
%\end{small}

\clearpage
\appendix
\section{Proofs} \label{sec:app_theory} %\subsection{Proofs of Theorems~\ref{thm:cost_hier} and \ref{thm:cost_flat}}
%\label{sec:mbproof}
\begin{proof}[Proof of Theorem~\ref{thm:cost_flat}]
The first term $T\CIfu$ should be obvious as the expert inspects the agent's overall behavior in each episode. Whenever something goes wrong in an episode, the expert labels the whole trajectory, incurring $\CLfu$ each time. The remaining work is to bound the number of episodes where agent makes one or more mistakes. This quantity is bounded by the number of total mistakes made by the halving algorithm, which is at most the logarithm of the number of candidate functions (policies), $\log |\Pifu| = \log\bigParens{|\Mcal| |\Pilo|^{|\Gcal|}} = \log |\Mcal| + |\Gcal| \log |\Pilo|$. This completes the proof.
\end{proof}

\begin{proof}[Proof of Theorem~\ref{thm:cost_hier}]
Similar to the proof of Theorem~\ref{thm:cost_flat}, the first term $T\CIfu$ is obvious. The second term corresponds to the situation where $\chkfu$ finds issues. According to Algorithm~\ref{alg:hierarchical_dagger}, the expert then labels the subgoals and also inspects whether each subgoal is accomplished successfully, which incurs $\CLhi + \Hhi \CIlo$ cost each time. The number of times that this situation happens is bounded by (a) the number of times that a wrong subgoal is chosen, plus (b) the number of times that all subgoals are good but at least one of the subpolicies fails to accomplish the subgoal. Situation (a) occurs at most $\log|\Mcal|$ times. In situation (b), the subgoals chosen in the episode must come from $\Gopt$, and for each of these subgoals the halving algorithm makes at most $\log|\Pilo|$ mistakes.
The last term corresponds to cost of $\lbllo$ operations. This only occurs when the meta-controller chooses a correct subgoal but the corresponding subpolicy fails. Similar to previous analysis, this situation occurs at most $\log|\Pilo|$ for each ``good'' subgoal ($g\in\Gopt$). This completes the proof.
\end{proof}

\section{Additional Experimental Details} \label{sec:app_experiment} %\newpage
%\section{Additional Experimental Details}
%\label{sec:app_experiment}
In our experiments, \emph{success rate} and \emph{external rewards} are reported as the trailing average over previous 100 episodes of training. For hierarchical imitation learning experiments in maze navigation domain, the success rate is only measured on separate test environments not used for training.

In addition to experimental results, in this section we describe our mechanism for subgoal detection / terminal predicate for Montezuma's Revenge and how the Maze Navigation environments are created. Network architectures from our experiments are in Tables~\ref{tab:arch_maze} and~\ref{tab:arch_montezuma}.

\subsection{Maze Navigation Domain}
\label{app:maze}

\begin{figure}
	\begin{center}
		\centerline{\includegraphics[width=0.8\columnwidth]{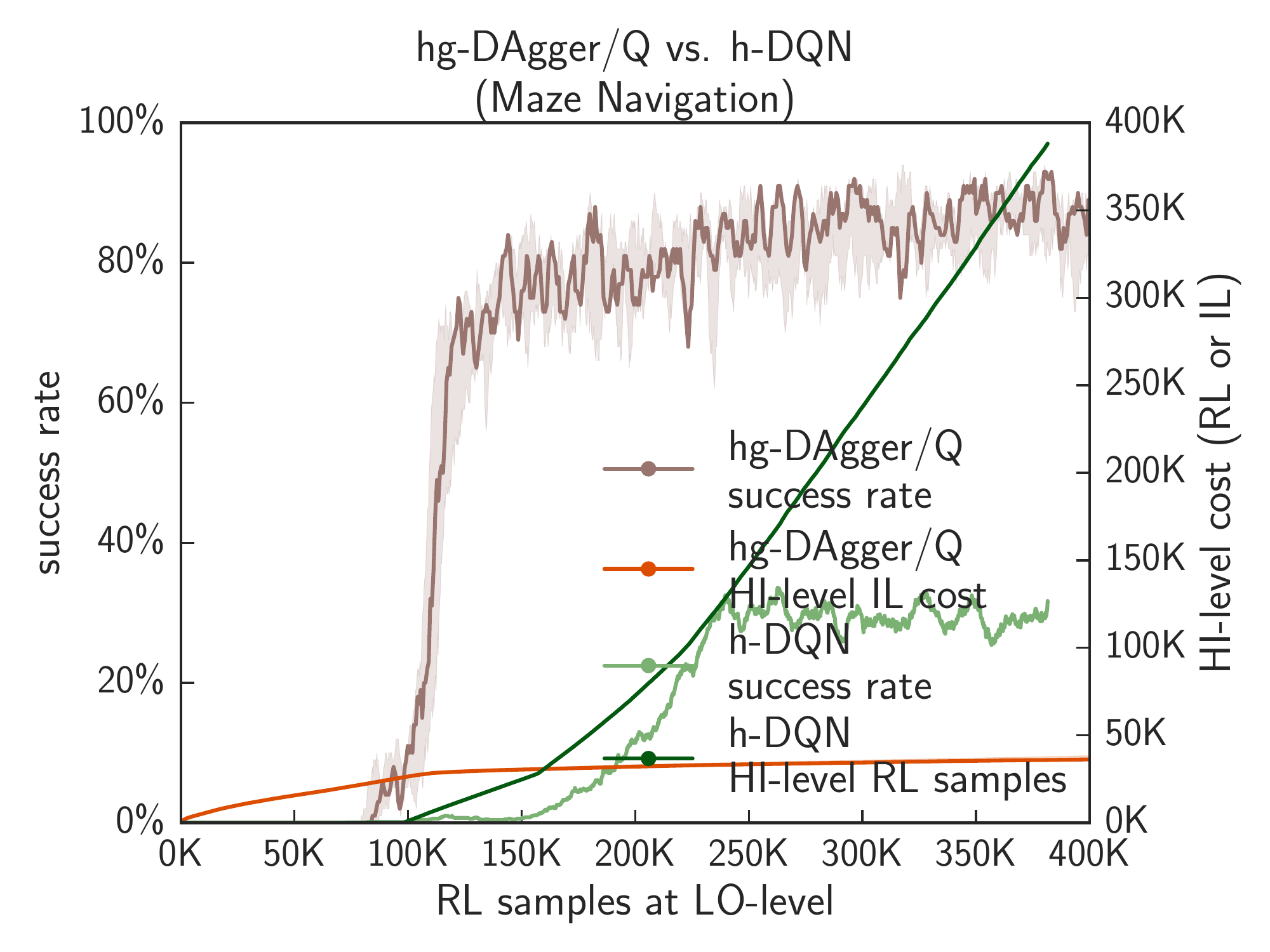}}
		\caption{\emph{Maze navigation: hybrid IL-RL (full task) versus h-DQN (with 50\% head-start).}}
		\label{fig:hybrid_vs_hdqn_maze}
	\end{center}
\end{figure}

We compare hg-DAgger/Q with the hierarchical reinforcement learning baseline (h-DQN, \citealp{kulkarni2016hierarchical}) with the same network architecture for the meta-controller and subpolicies as hg-DAgger/Q and similarly enhanced $Q$-learning procedure.

Similar to the Montezuma's Revenge domain, h-DQN does not work well for the maze domain.
At the \hi level, the planning horizon of 10--12 with 4--5 possible subgoals in each step is prohibitively difficult for the $\hi$-level reinforcement learner and we were not able to achieve non-zero rewards within in any of our experiments.
To make the comparison, we attempted to provide additional advantage to the h-DQN algorithm by giving it some head-start, so we ran h-DQN with 50\% reduction in the horizon, by giving the hierarchical learner the optimal execution of the first half of the trajectory. The resulting success rate is in Figure~\ref{fig:hybrid_vs_hdqn_maze}. Note that the hybrid IL-RL does not get the 50\% advantage, but it still quickly outperforms h-DQN, which flattens out at 30\% success rate.

\begin{figure*}[t]
	\centering     %%% not \center
	\customlabel{fig:screenshot2}{fig:maze_extra}{shot2}%
	\includegraphics[width=0.24\textwidth]{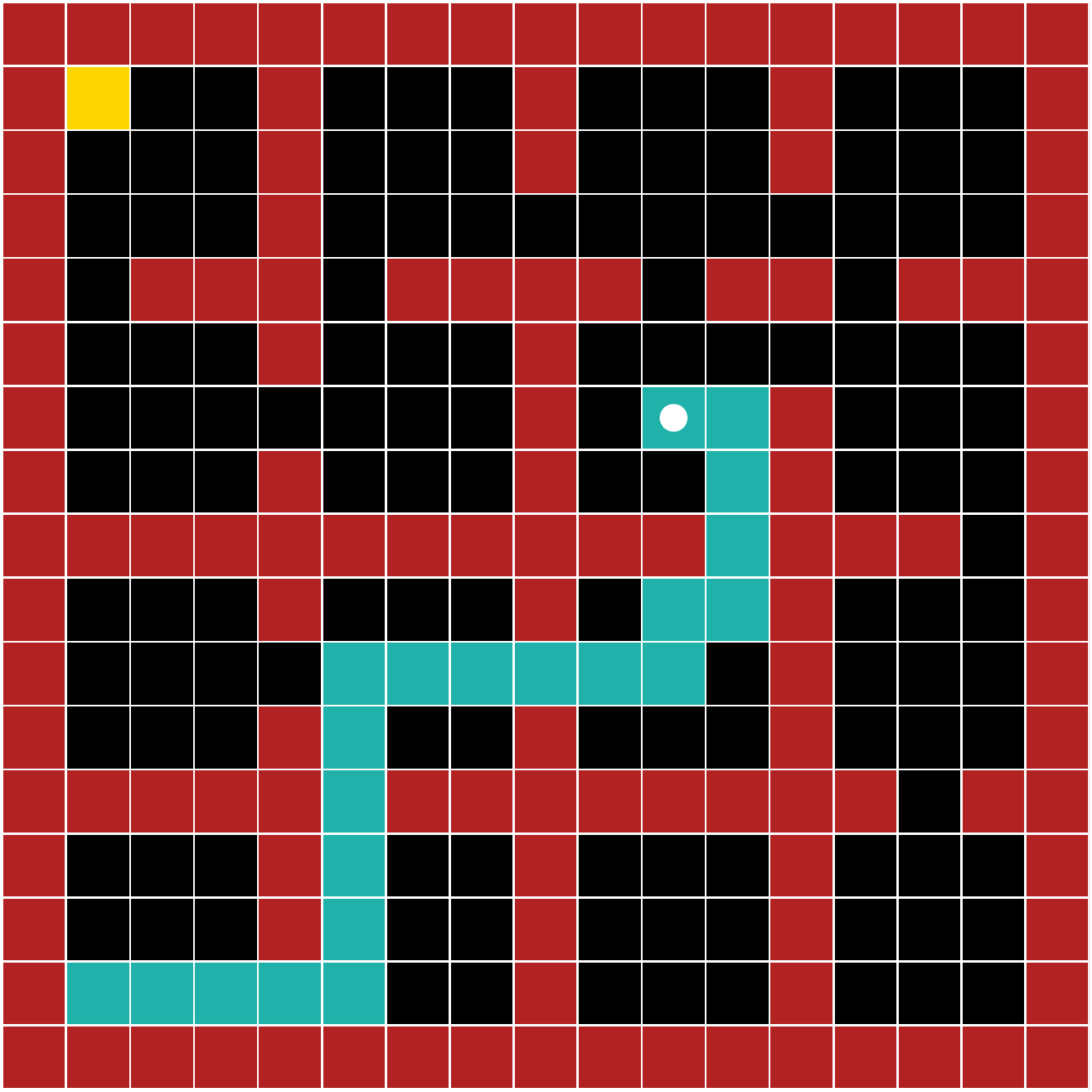}%
	\hfill%
	\customlabel{fig:screenshot3}{fig:maze_extra}{shot3}%
	\includegraphics[width=0.24\textwidth]{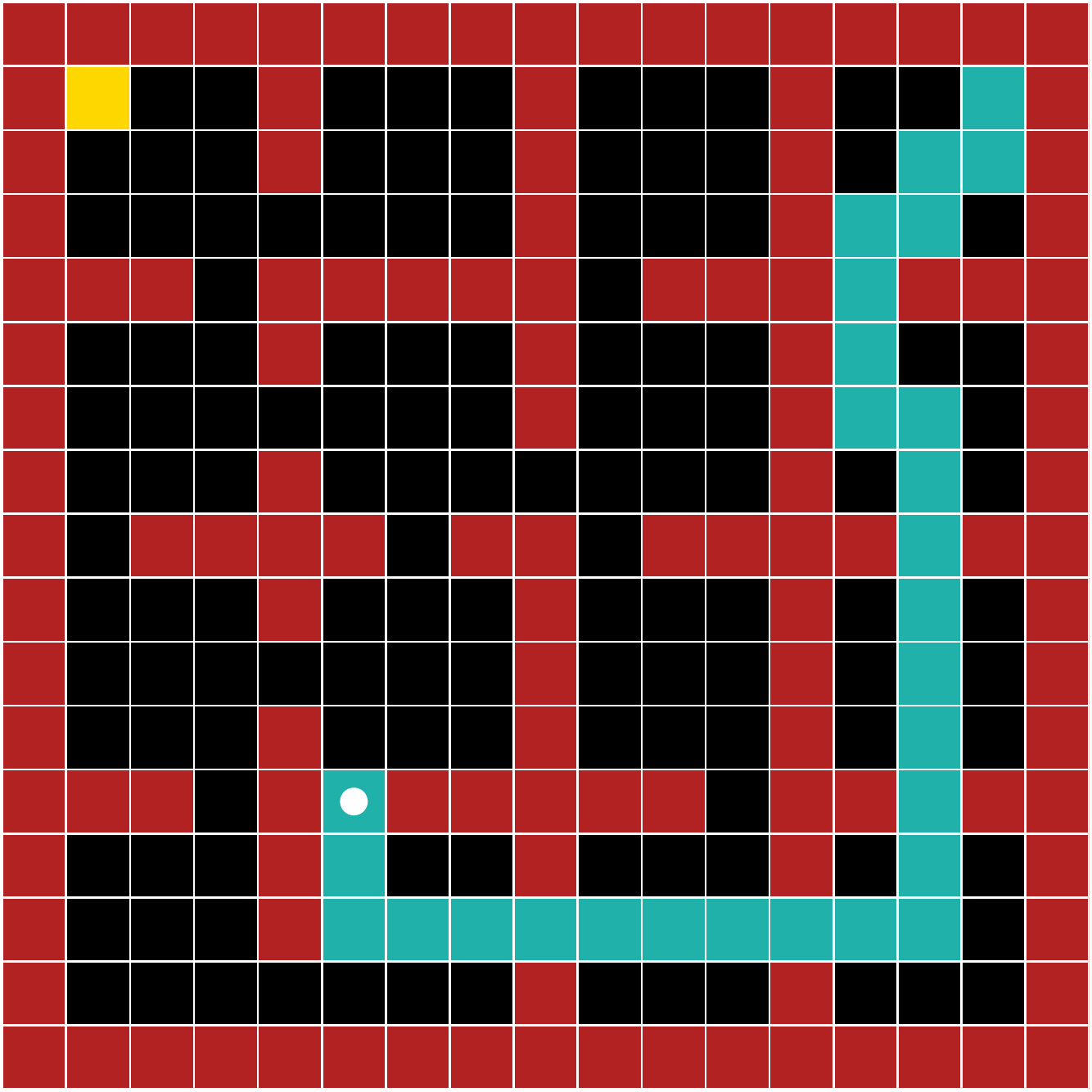}%
	\hfill%
	\customlabel{fig:screenshot4}{fig:maze_extra}{shot4}%
	\includegraphics[width=0.24\textwidth]{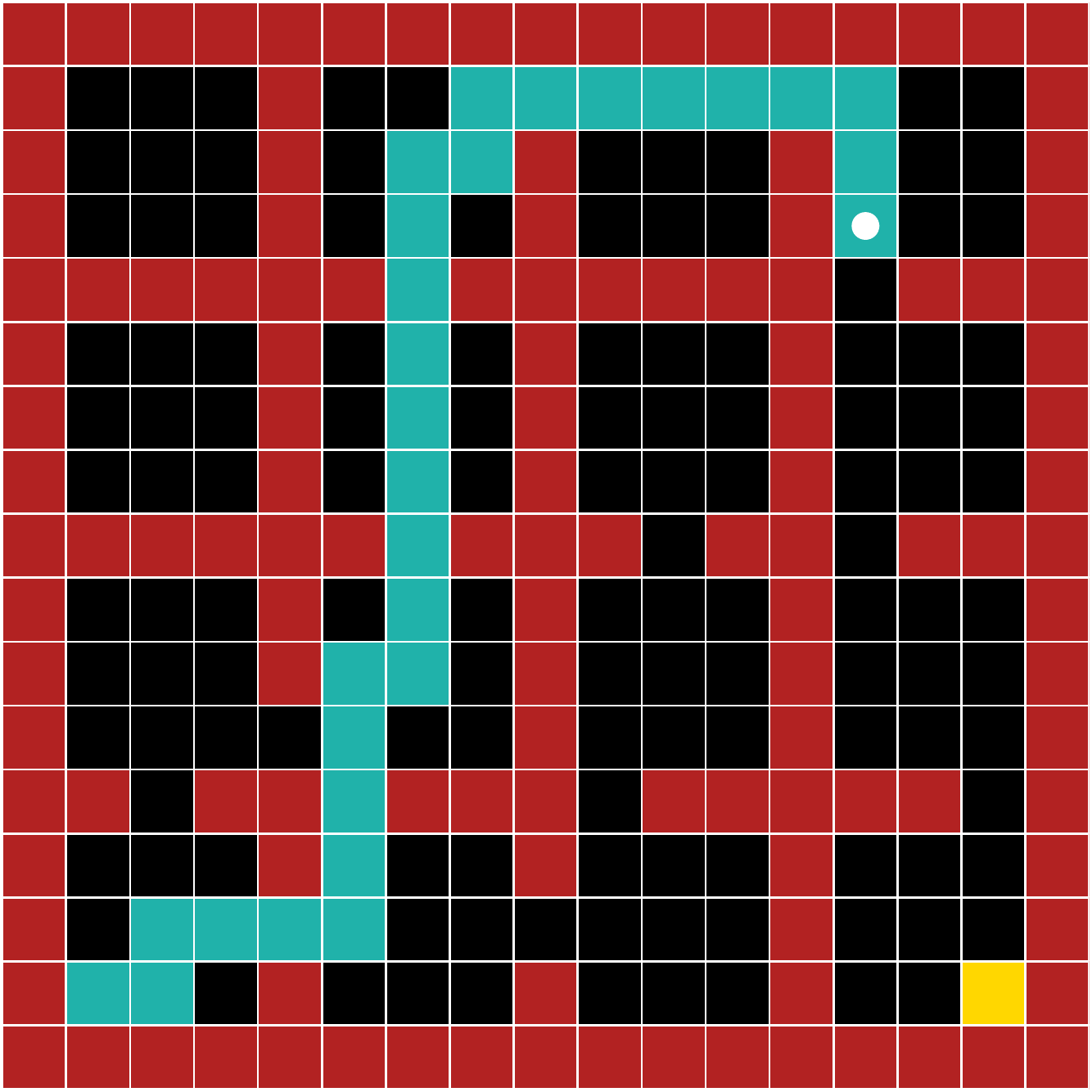}%
	\hfill%
	\customlabel{fig:screenshot5}{fig:maze_extra}{shot5}%
	\includegraphics[width=0.24\textwidth]{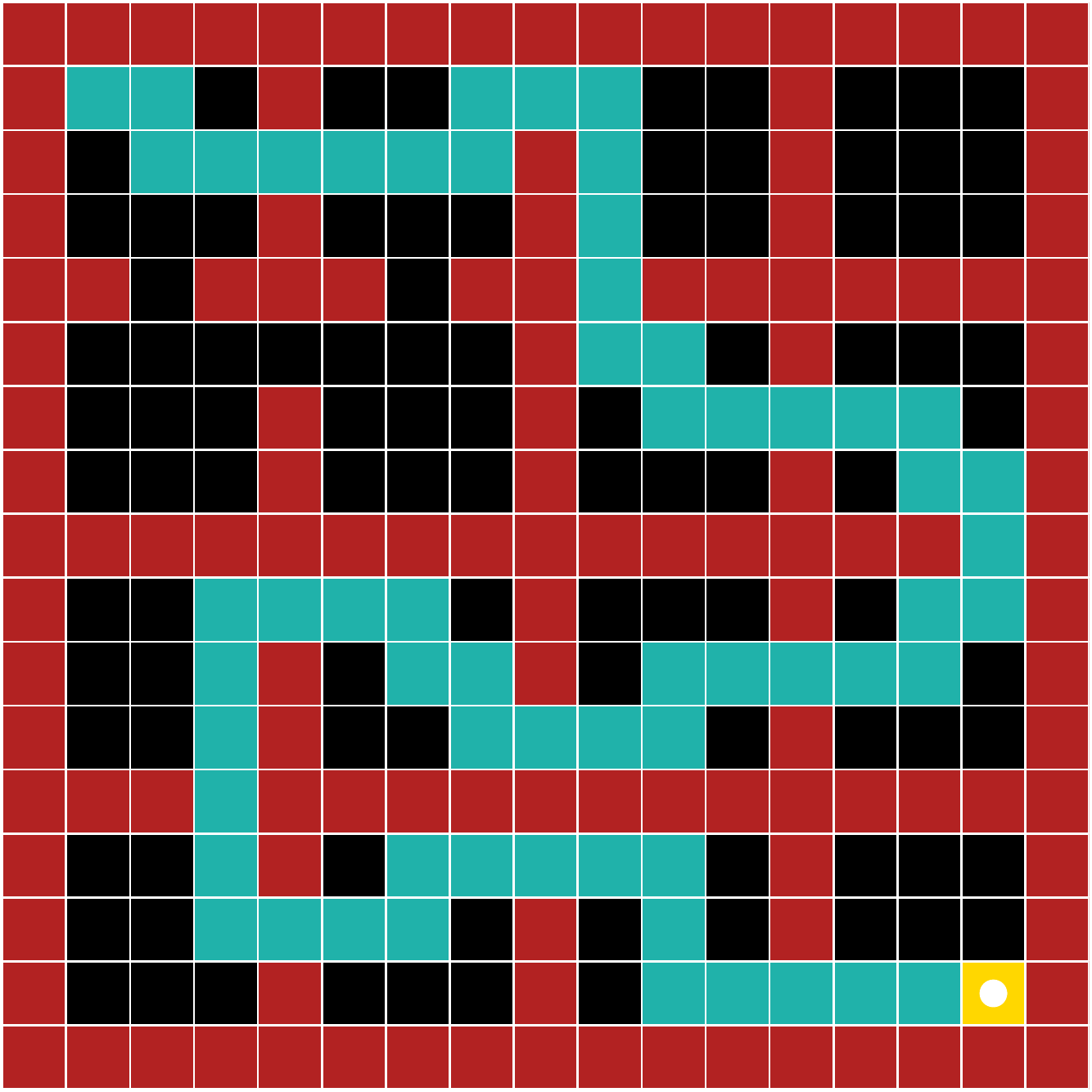}%
	\vspace{-0.1in}
	\caption{\emph{Maze navigation.} Sample random instances of the maze domain (different from main text). The $17\times 17$ pixel representation of the maze is used as input for neural network policies.}
	\label{fig:another_maze_domain}
\end{figure*}
\subsubsection{Creating Maze Navigation Environments}
We create 2000 maze navigation environments, 1000 of which are used for training and 1000 maps are used for testing. The comparison results for maze navigation (e.g., Figure~\ref{fig:cmp:flat}) are all based on randomly selected environments among 1000 test maps. See Figure~\ref{fig:another_maze_domain} for additional examples of the environments created. For each map (environment instance), we start with a 17$\times$17 grid, which are divided into 4$\times$4 room structure. Initially, no door exists in between rooms. To create an instance of the maze navigation environment, the goal block (yellow) and the starting position are randomly selected (accepted as long as they are not the same). Next, we randomly select a wall separating two different room and replace a random red block (lava) along this wall with a door (black cell). This process continues until two conditions are satisfied:
\begin{itemize}
	\item There is a feasible path between the starting location and the goal block (yellow)
	\item The minimum distance between start to goal is at least 40 steps. The optimal path can be constructed using graph search
\end{itemize}
Each of the 2000 environments create must satisfy both conditions. The expert labels for each environment come from optimal policy computed via value iteration (which is fast based on tabular representation of the given grid world).
\subsubsection{Hyperparameters for Maze Navigation}
The network architecture used for maze navigation is described in Table~\ref{tab:arch_maze}. The only difference between subgoal policy networks and metacontroller network is the number of output class (4 actions versus 5 subgoals). For our hierarchical imitation learning algorithms, we also maintain a small network along each subgoal policy for subgoal termination classification (one can also view the subgoal termination classifier as an extra head of the subgoal policy network).

The contextual input (state) to the policy networks consists of 3-channel pixel representation of the maze environment. We assign different (fixed) values to goal block, agent location, agent's trail and lava blocks. In our hierarchical imitation learning implementations, the base policy learner (DAgger and behavior cloning) update the policies every 100 steps using stochastic optimization. We use Adam optimizer and learning rate of 0.0005.
\begin{table}[h]\caption{Network Architecture---Maze Domain}%
	\smallskip%
	\small%
	\begin{center}% used the environment to augment the vertical space
		% between the caption and the table
		\begin{tabular}{ll}
			\toprule
			1: Convolutional Layer & 32 filters, kernel size 3, stride 1\\
			2: Convolutional Layer & 32 filters, kernel size 3, stride 1\\
			3: Max Pooling Layer & pool size $2\time 2$\\[3pt]
			4: Convolutional Layer  & 64 filters, kernel size 3, stride 1\\
			5: Convolutional Layer & 64 filters, kernel size 3, stride 1\\
			6: Max Pooling Layer & pool size $2\time 2$\\[3pt]
			7: Fully Connected Layer & 256 nodes, relu activation\\[3pt]
			8: Output Layer & softmax activation\\
			& (dimension 4 for subpolicy,\\
			& dimension 5 for meta-controller)\\
			\bottomrule
		\end{tabular}
	\end{center}
	\label{tab:arch_maze}
\end{table}

\subsection{Montezuma's Revenge}
\begin{figure*}
	\centering     %%% not \center
	\customlabel{fig:first}{fig:montezuma_screenshot}{first}%
	\includegraphics[width=0.24\textwidth]{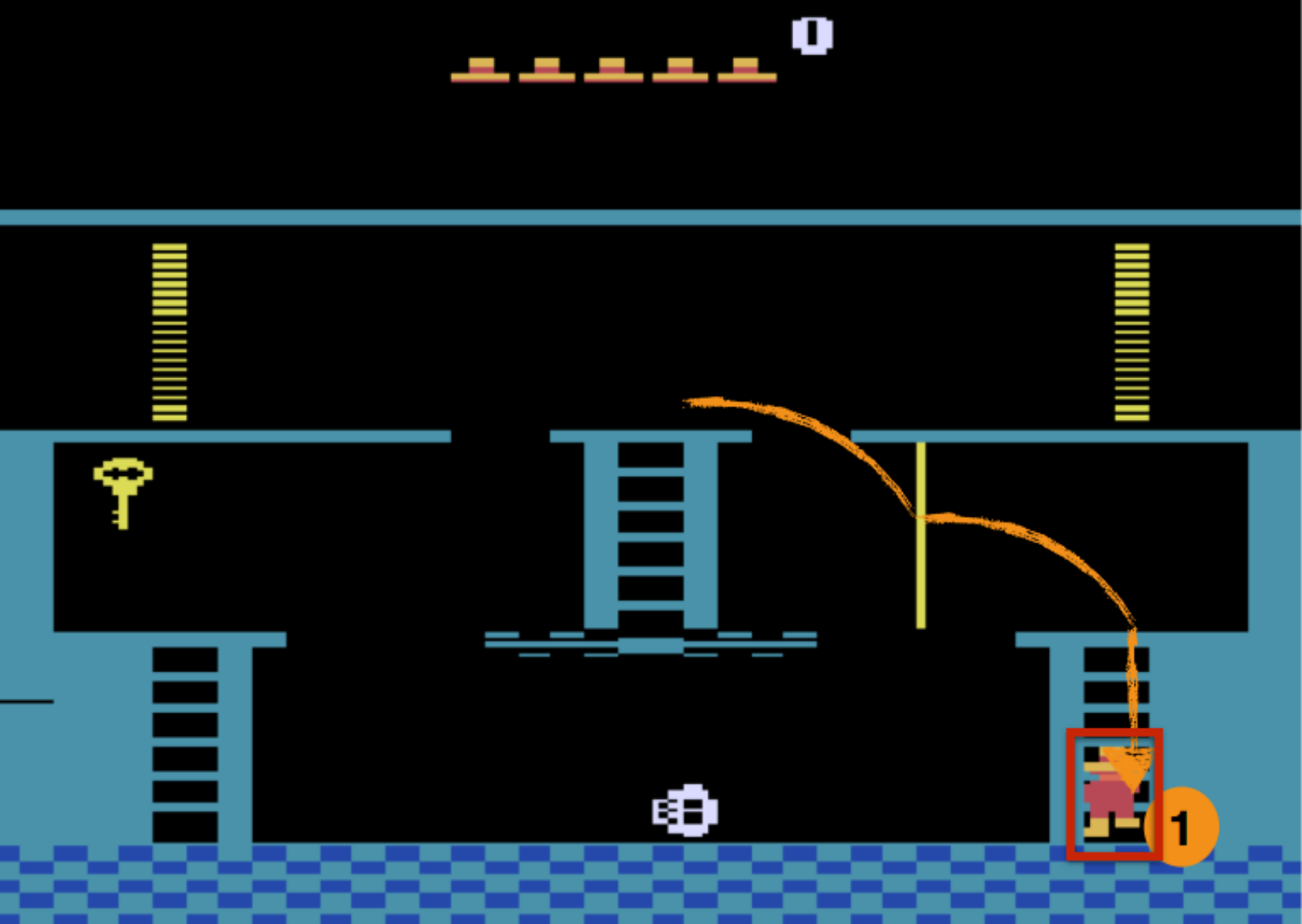}%
	\hfill%
	\customlabel{fig:second}{fig:montezuma_screenshot}{second}%
	\includegraphics[width=0.24\textwidth]{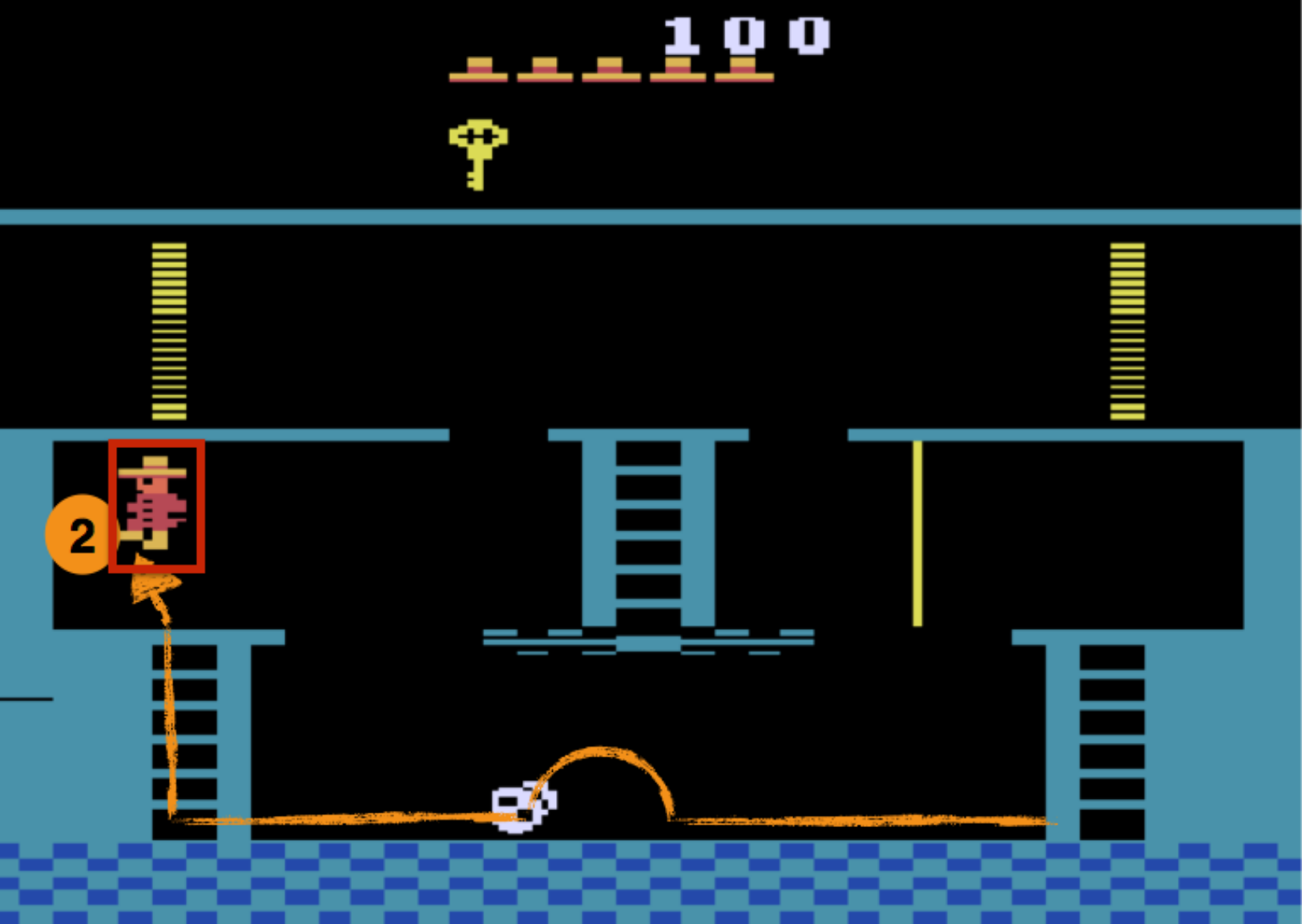}%
	\hfill%
	\customlabel{fig:third}{fig:montezuma_screenshot}{third}%
	\includegraphics[width=0.24\textwidth]{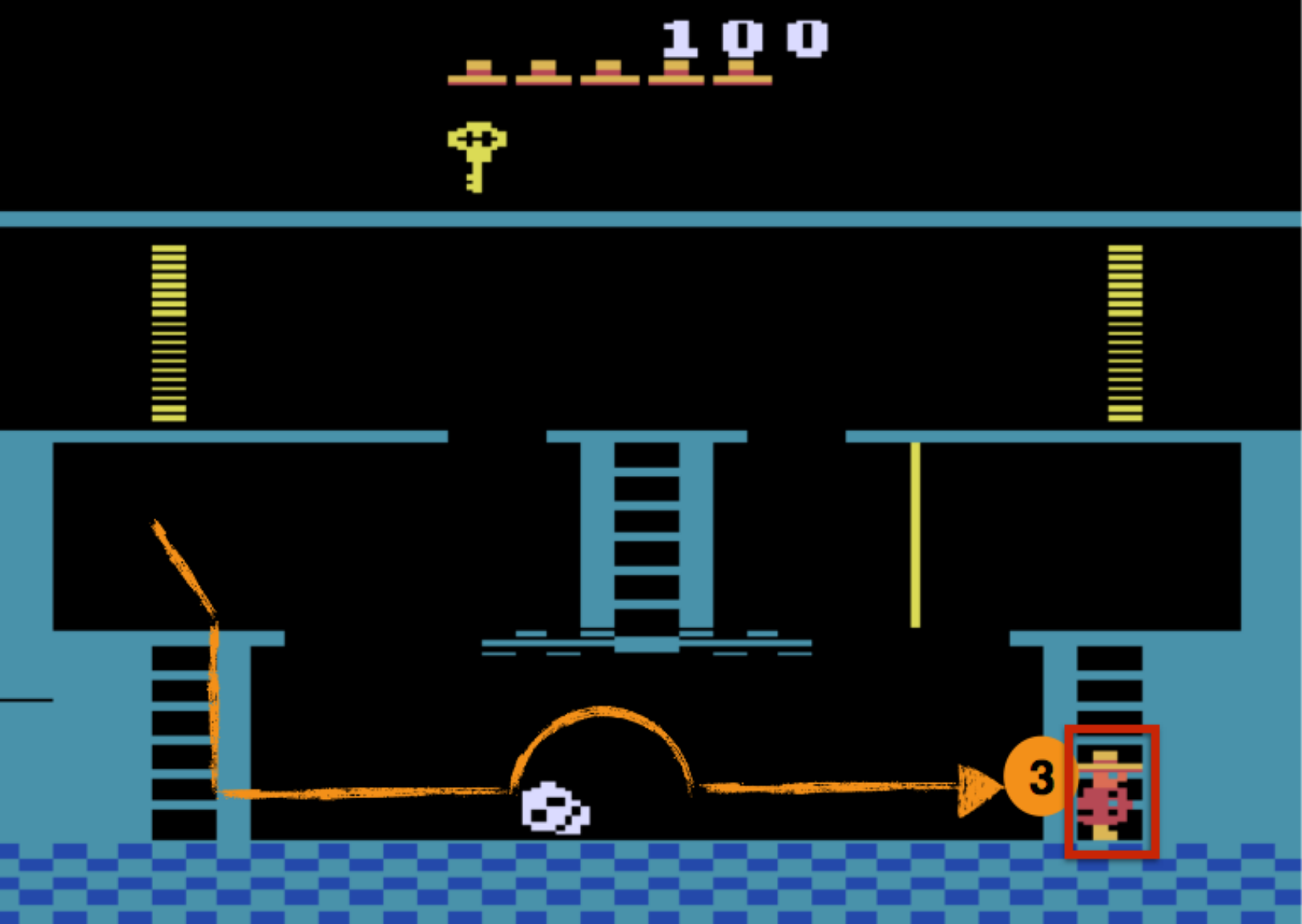}%
	\hfill%
	\customlabel{fig:fourth}{fig:montezuma_screenshot}{fourth}%
	\includegraphics[width=0.24\textwidth]{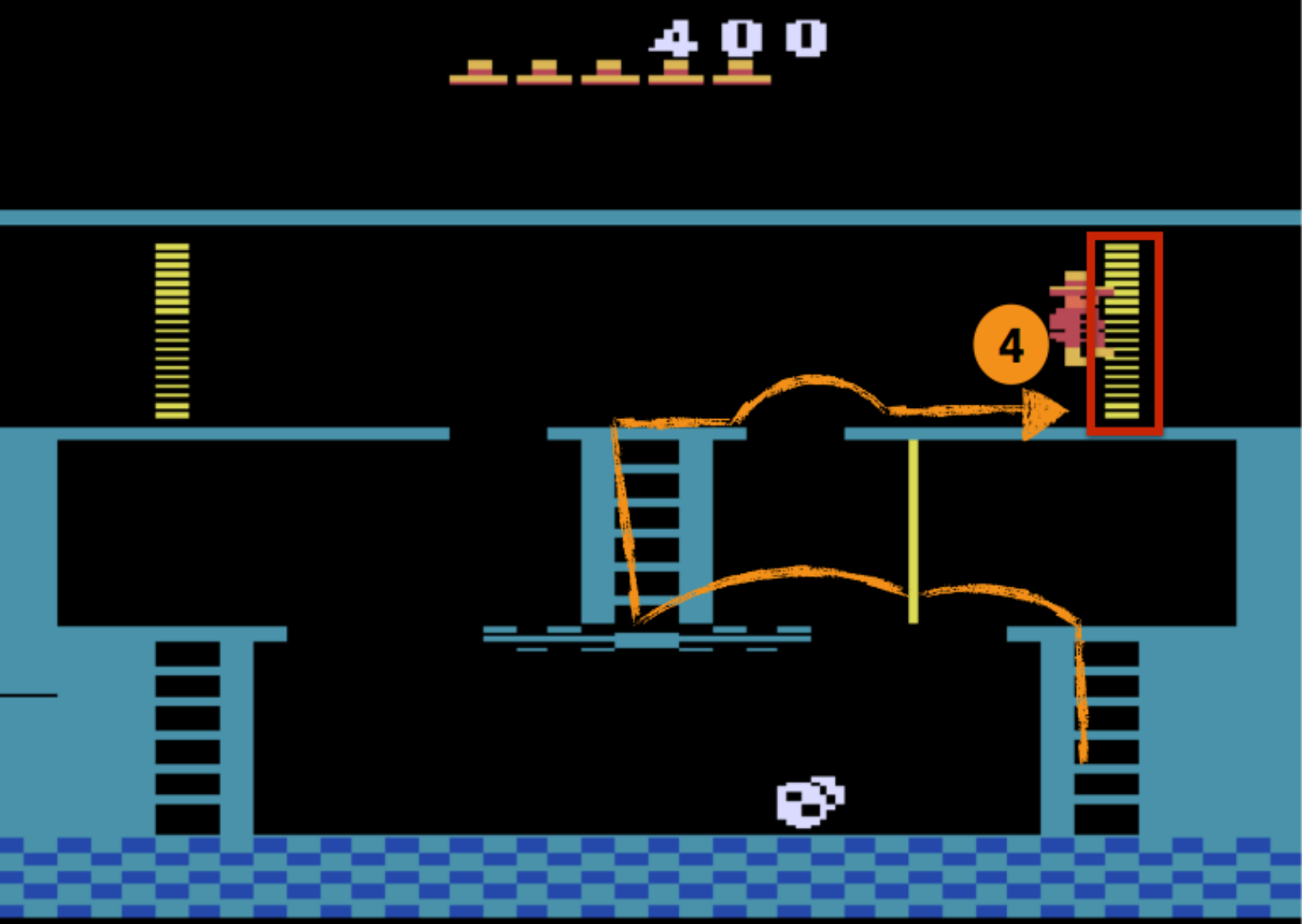}%
	\vspace{-0.1in}
	\caption{\emph{Montezuma's Revenge: Screenshots of the environment with 4 designated subgoals in sequence.}}
	\label{fig:montezuma_4_subgoals}
\end{figure*}

\begin{figure*}
	\centering
	\customlabel{fig:montezuma_best10}{fig:montezuma_app}{left}%
	\includegraphics[width=0.33\textwidth]{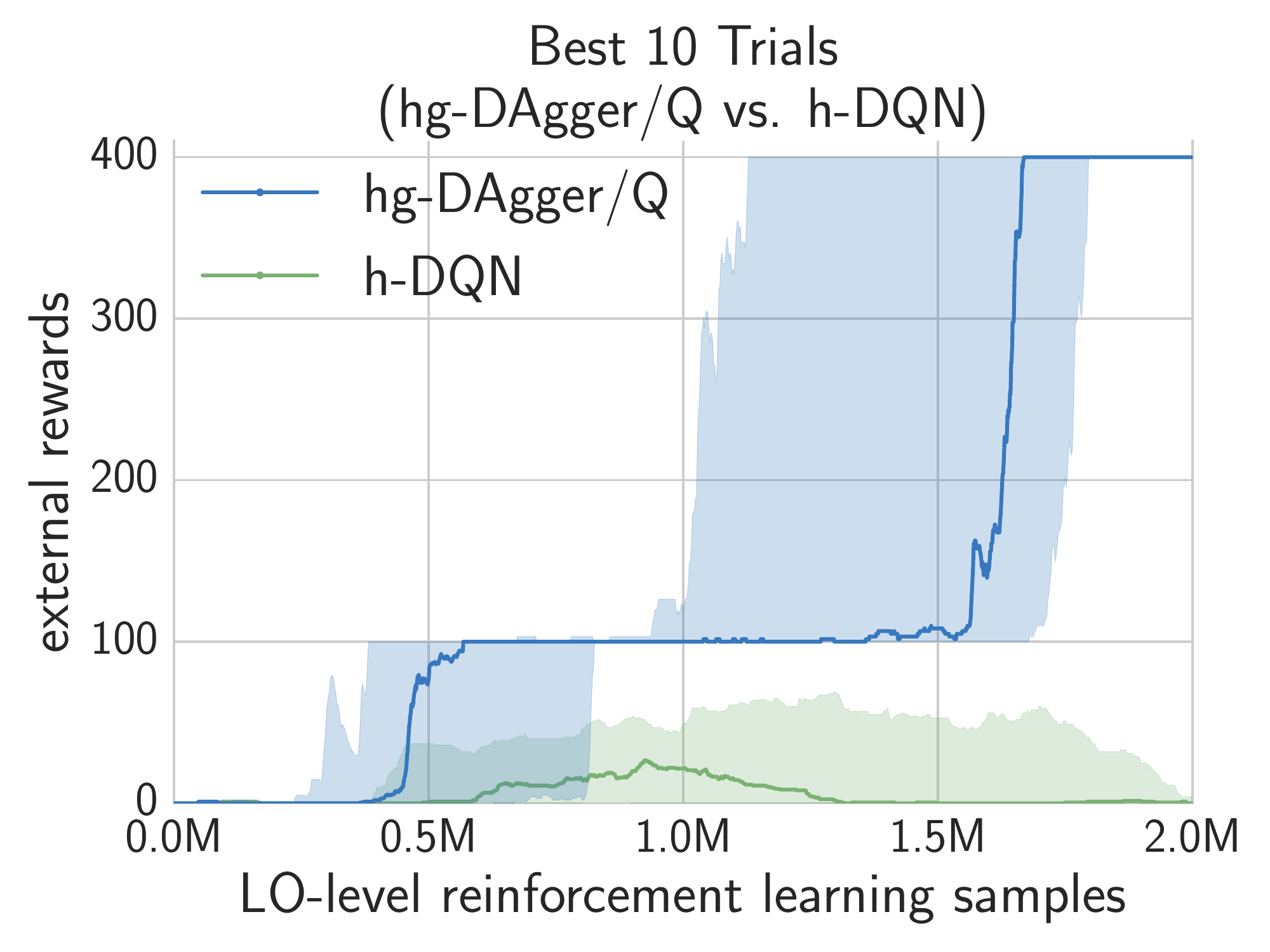}%
	\hfill%
	\customlabel{fig:atari_subgoals_completion}{fig:montezuma_app}{middle}%
	\includegraphics[width=0.33\textwidth]{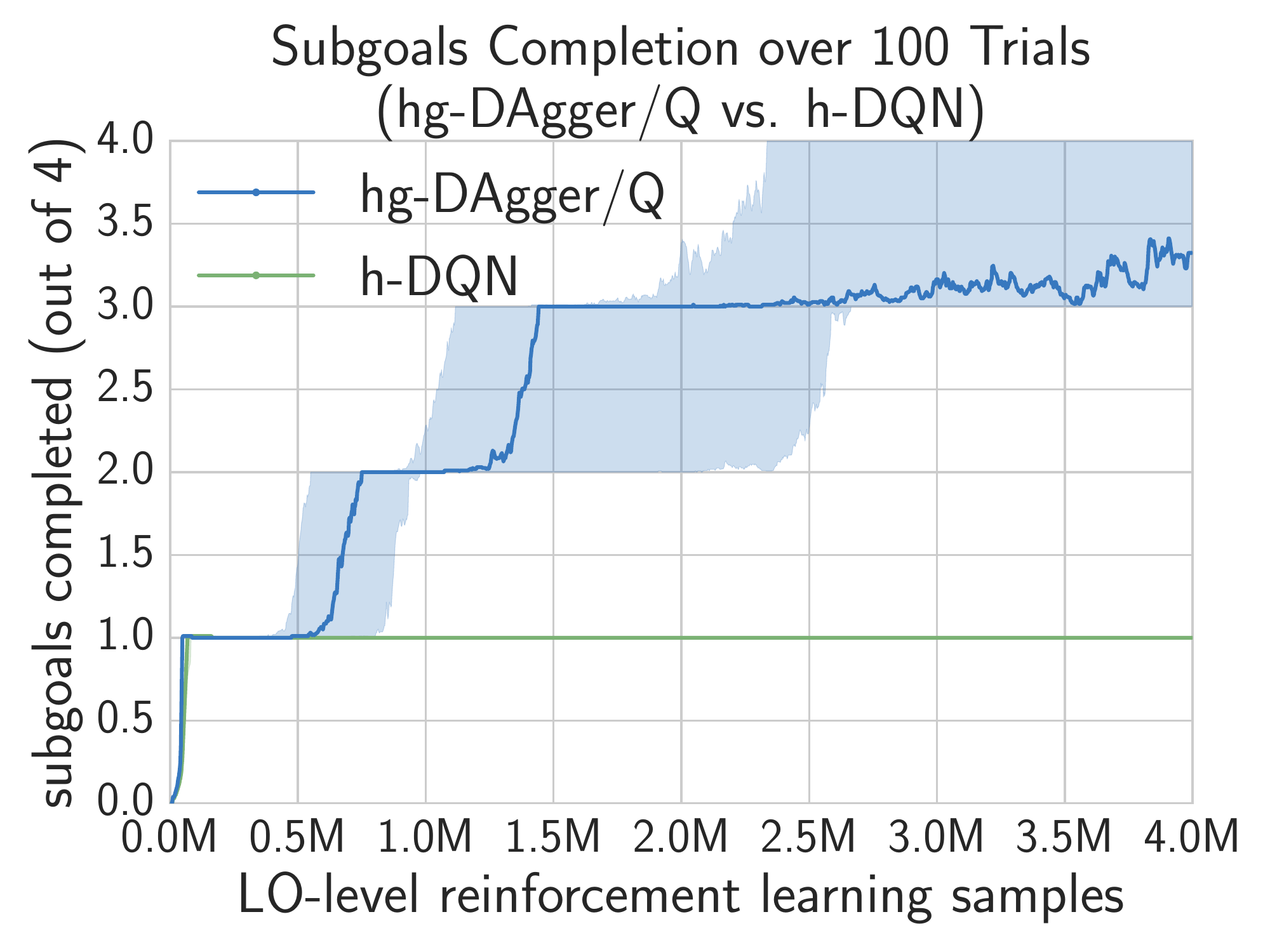}%
	\hfill%
	\customlabel{fig:atari_key}{fig:montezuma_app}{right}%
	\includegraphics[width=0.33\textwidth]{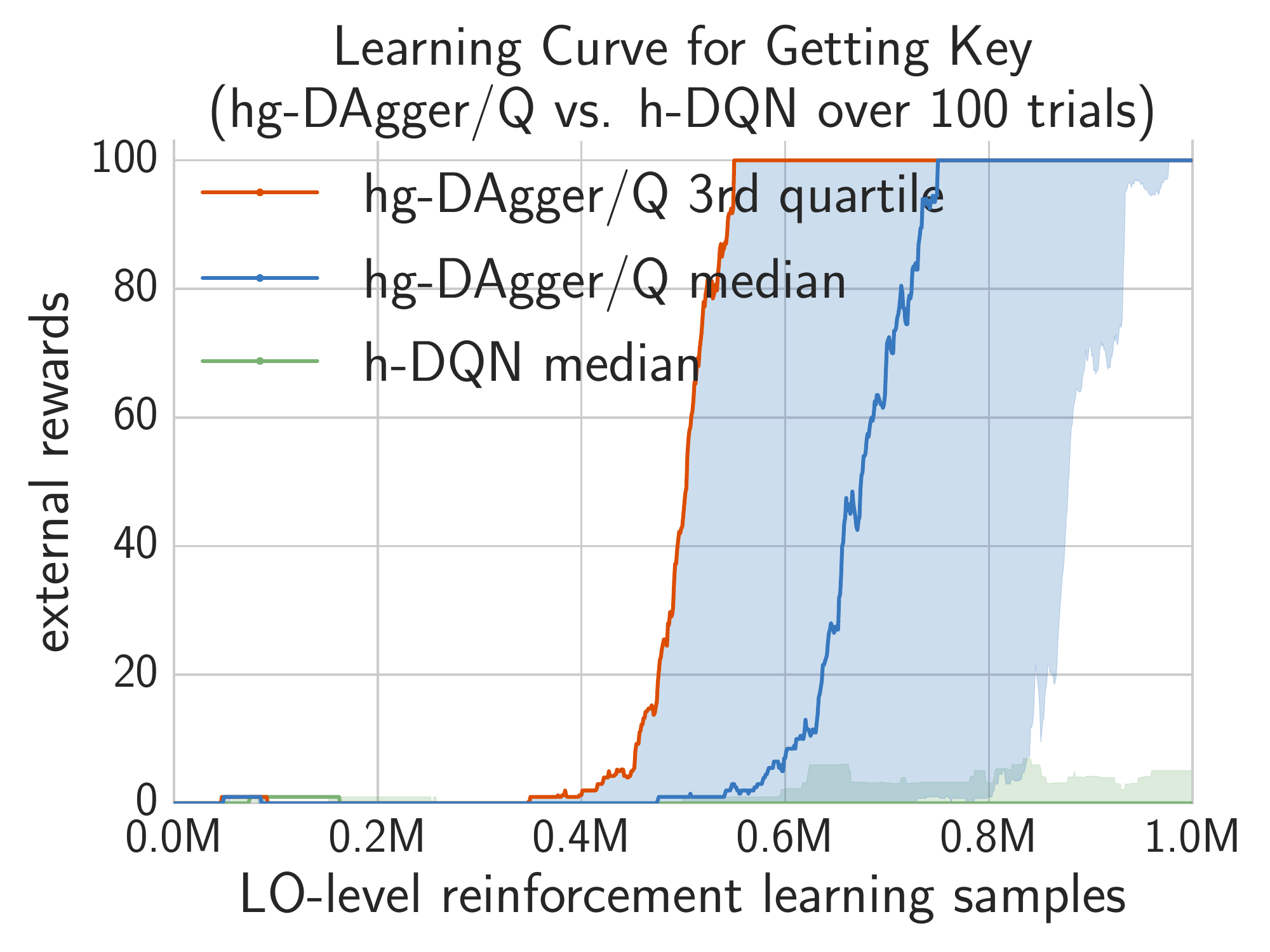}%
	\vspace{-0.1in}
	\caption{\emph{Montezuma's revenge: hybrid IL-RL versus hierarchical RL.}
		\emph{(Left)} Median reward, min and max across the best 10 trials. The agent
		completes the first room in less than 2 million samples. The shaded region
		corresponds to min and max of the best 10 trials.
		\emph{(Middle)} Median, first and third quartile
		of subgoal completion rate across 100 trials. The shaded region
		corresponds to first and third quartile.
		\emph{(Right)} Median, first and third quartile of reward
		across 100 trials. The shaded region corresponds to first and third
		quartile. h-DQN only considers the first two subgoals to simplify
		the learning task.}
	\label{fig:montezuma_app}
\end{figure*}

\begin{figure*}[ht!]
	\centering     %%% not \center
	\customlabel{fig:progress2}{fig:learning_progression}{progress2}%
	\includegraphics[width=0.33\textwidth]{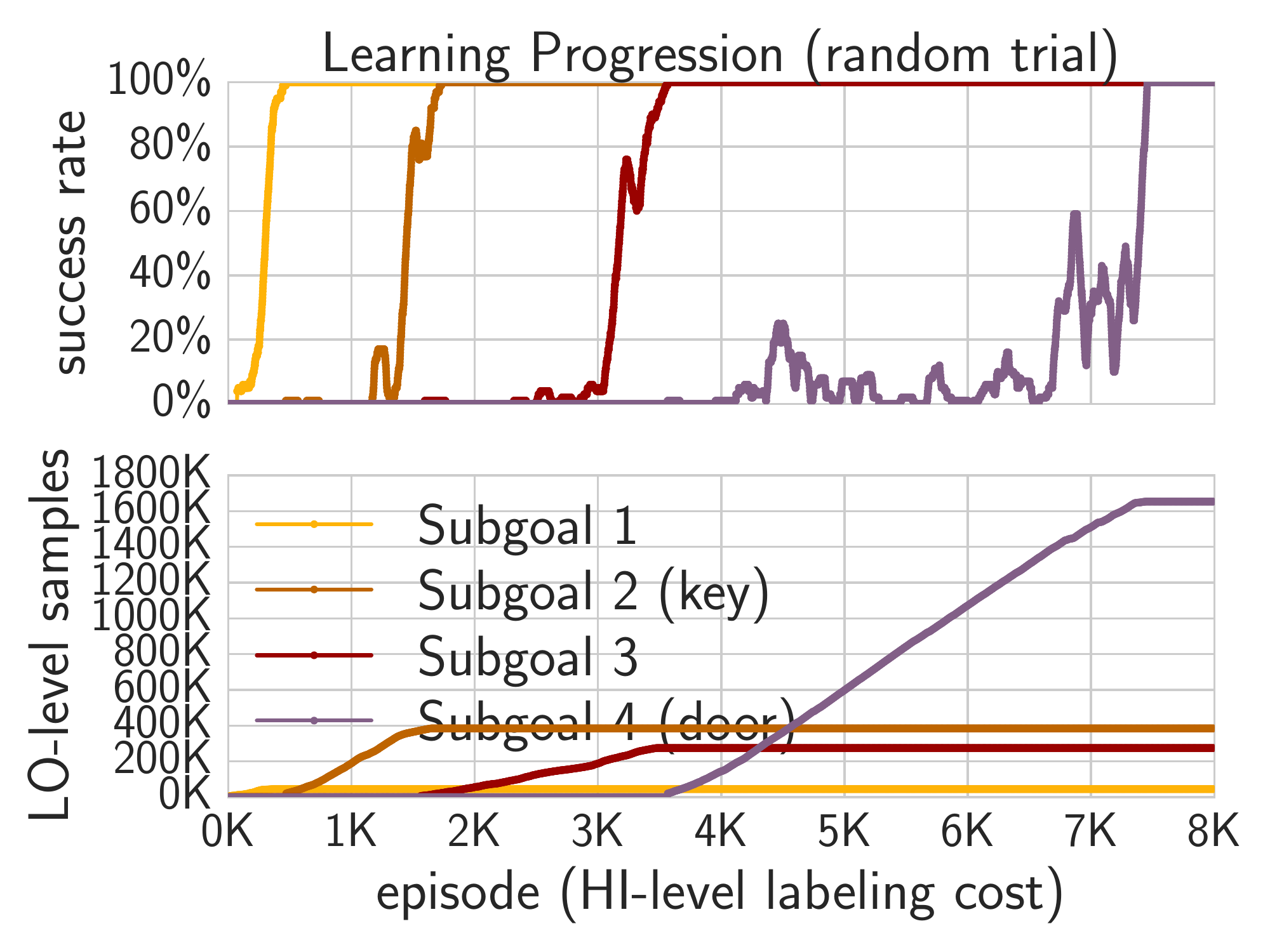}%
	\hfill%
	\customlabel{fig:progress3}{fig:learning_progression}{progress3}%
	\includegraphics[width=0.33\textwidth]{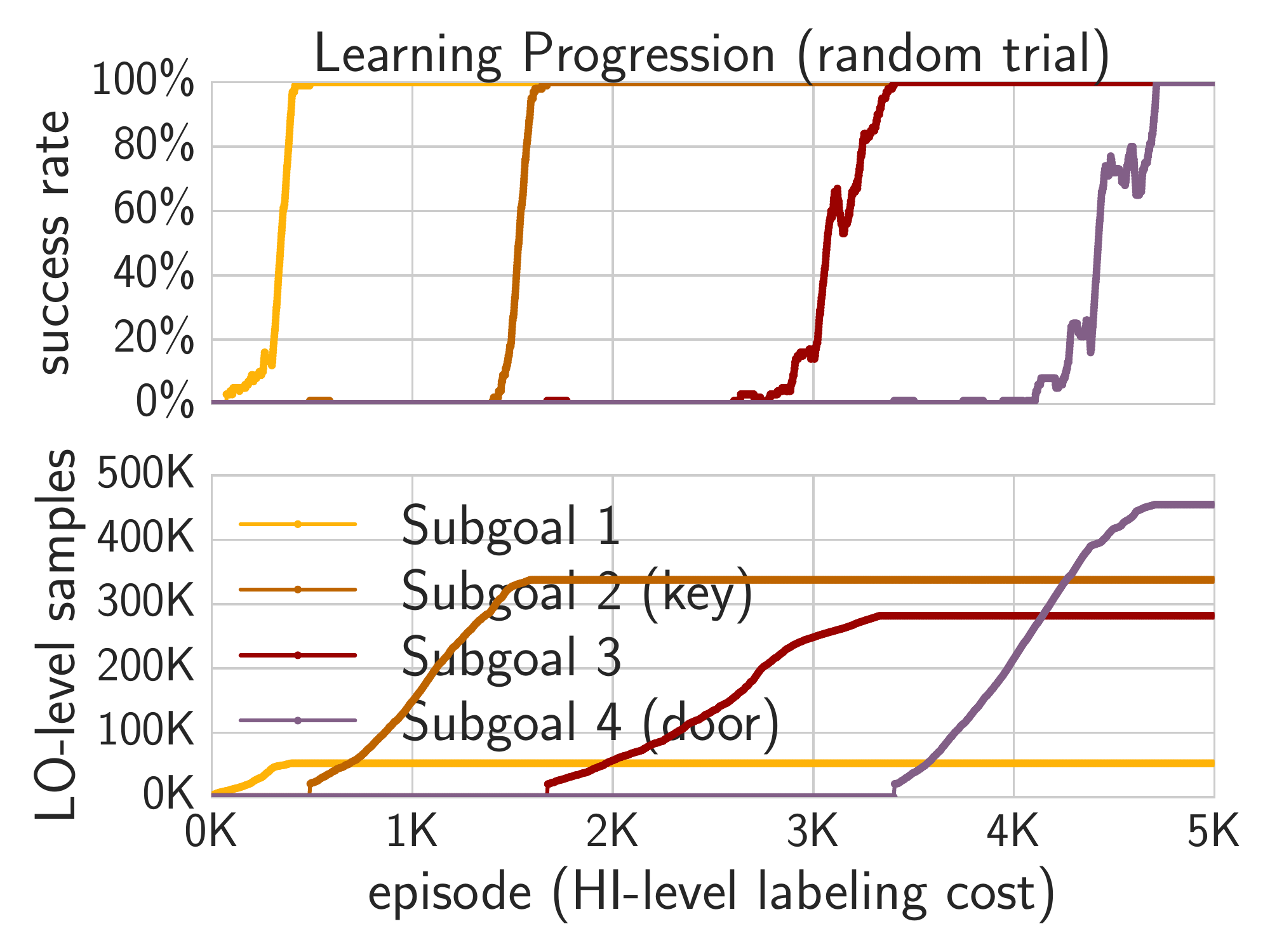}%
	\hfill%
	\customlabel{fig:progress4}{fig:learning_progression}{progress4}%
	\includegraphics[width=0.33\textwidth]{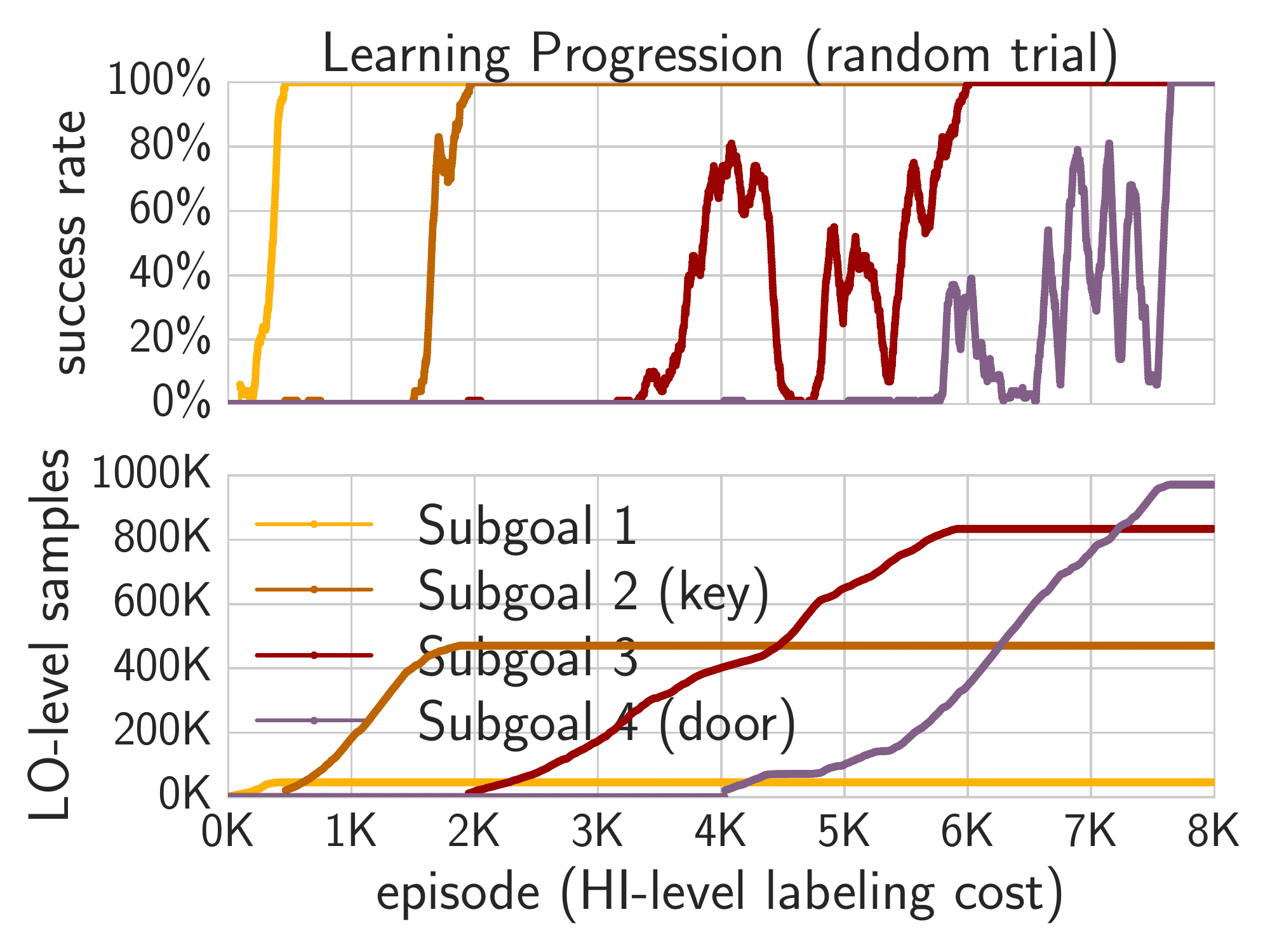}%
	\vspace{-0.1in}
	\caption{\emph{Montezuma's revenge: Learning progression of Algorithm 3 in solving the entire first room. The figures show three randomly selected successful trials.}}
	\label{fig:montezuma_learning_progression_extra}
\end{figure*}

\begin{figure}
	\begin{center}
		\centerline{\includegraphics[width=0.8\columnwidth]{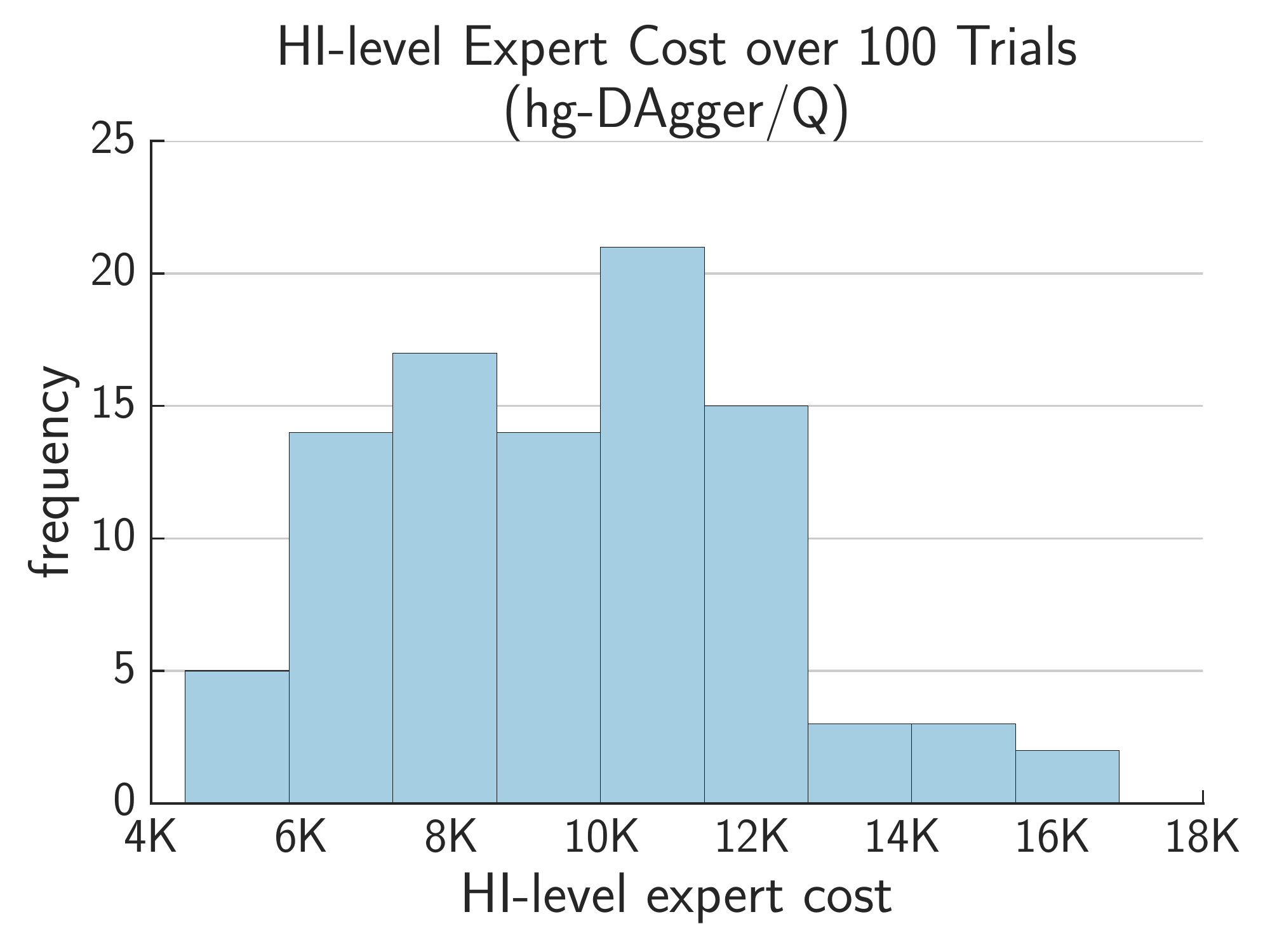}}
		\caption{\emph{Montezuma's Revenge: Number of HI-level expert labels.} Distribution of Hi-level expert labels needed across 100 trials; the histogram excludes 6 outliers whose number of labels exceeds 20K for ease of visualization}
		\label{fig:hybrid_vs_hdqn_meta_labels_hist}
	\end{center}
\end{figure}

Although the imitation learning component tends to be stable and consistent, the samples required by the reinforcement learners can vary between experiments with identical hyperparameters. In this section, we report additional results of our hybrid algorithm for the Montezuma's Revenge domain.

For the implementation of our hybrid algorithm on the game Montezuma's Revenge, we decided to limit the computation to 4 million frames for the \lo-level reinforcement learners (in aggregate across all 4 subpolicies). Out of 100 experiments, 81 out of 100 successfully learn the first 3 subpolicies, 89 out of 100 successfully learn the first 2 subpolicies. The last subgoal (going from the bottom of the stairs to open the door) proved to be the most difficult and almost half of our experiments did not manage to finish learning the fourth subpolicy within the 4 million frame limit (see Figure~\ref{fig:montezuma_app} middle pane). The reason mainly has to do with the longer horizon of subgoal 4 compared to other three subgoals. Of course, this is a function of the design of subgoals and one can always try to shorten the horizon by introducing intermediate subgoals.

However, it is worth pointing out that even as we limit the h-DQN baseline to only 2 subgoals (up to getting the key), the h-DQN baseline generally tends to underperform our proposed hybrid algorithm by a large margin. Even with the given advantage we confer to our implementation of h-DQN, all of the h-DQN experiments failed to successfully master the second subgoal (getting the key). It is instructive to also examine the sample complexity associated with getting the key (the first positive external reward, see Figure~\ref{fig:montezuma_app} right pane). Here the horizon is sufficiently short to appreciate the difference between having expert feedback at the $\hi$ level versus relying only on reinforcement learning to train the meta-controller.

The stark difference in learning performance (see Figure~\ref{fig:montezuma_app} right) comes from the fact that the $\hi$-level expert advice effectively prevents the \lo-level reinforcement learners from accumulating bad experience, which is frequently the case for h-DQN. The potential corruption of experience replay buffer also implies at in our considered setting, learning with hierarchical DQN is no easier compared to flat DQN learning. Hierarchical DQN is thus susceptible to collapsing into the flat learning version.

\subsubsection{Subgoal Detectors for Montezuma's Revenge}
In principle, the system designer would select the hierarchical decomposition that is most convenient for giving feedback. For Montezuma's Revenge, we set up four subgoals and automatic detectors that make expert feedback trivial. The subgoals are landmarks that are described by small rectangles. For example, the door subgoal (subgoal 4) would be represented by a patch of pixel around the right door (see Figure~\ref{fig:montezuma_4_subgoals} right). We can detect the correct termination / attainment of this subgoal by simply counting the number of pixels inside of the pre-specified box that has changed in value. Specifically in our case, subgoal completion is detected if at least 30\% of pixels in the landmark's detector box changes.

\subsubsection{Hyperparameters for Montezuma's Revenge}
\begin{table}[h]\caption{Network Architecture---Montezuma's Revenge}%
	\smallskip%
	\small%
	\begin{center}% used the environment to augment the vertical space
		% between the caption and the table
		\begin{tabular}{ll}
			\toprule
			1: Conv.\ Layer & 32 filters, kernel size 8, stride 4, relu\\
			2: Conv.\ Layer & 64 filters, kernel size 4, stride 2, relu\\
			3: Conv.\ Layer  & 64 filters, kernel size 3, stride 1, relu\\[3pt]
			4: Fully Connected & 512 nodes, relu,\\
			\hphantom{4: }%
			Layer           & normal initialization with std 0.01\\[3pt]
			5: Output Layer & linear (dimension 8 for subpolicy,\\
			& dimension 4 for meta-controller)\\
			\bottomrule
		\end{tabular}
	\end{center}
	\label{tab:arch_montezuma}
\end{table}

Neural network architecture used is similar to \cite{kulkarni2016hierarchical}. One difference is that we train a separate neural network for each subgoal policy, instead of maintaining a subgoal encoding as part of the input into a policy neural network that shares representation for multiple subgoals jointly. Empirically, sharing representation across multiple subgoals causes the policy performance to degrade we move from one learned subgoal to the next (a phenomenon of catastrophic forgetting in deep learning literature). Maintaining each separate neural network for each subgoal ensures the performance to be stable across subgoal sequence. The metacontroller policy network also has similar architecture. The only difference is the number of output (4 output classes for metacontroller, versus 8 classes (actions) for each \lo-level policy).

For training the \lo-level policy with $Q$-learning, we use DDQN \cite{van2016deep} with prioritized experience replay \cite{schaul2015prioritized} (with prioritization exponent $\alpha=0.6$, importance sampling exponent $\beta_0 = 0.4$). Similar to previous deep reinforcement learning work applied on Atari games, the contextual input (state) consists of four consecutive frames, each converted to grey scale and reduced to size 84$\times$ 84 pixels. Frame skip parameter as part of the Arcade Learning Environment is set to the default value of 4. The repeated action probability is set to 0, thus the Atari environment is largely deterministic. The experience memory has capacity of 500K. The target network used in $Q$-learning is updated every 2000 steps. For stochastic optimization, we use rmsProp with learning rate of 0.0001, with mini-batch size of 128.

\section{Additional Related Work} \label{sec:app_related} %\subsection{Other Related Works}
\label{sec:others}
\textbf{Imitation Learning.} Another dichotomy in imitation learning, as well as in reinforcement learning, is that of value-function learning versus policy learning.  The former setting  \cite{abbeel2004apprenticeship,ziebart2008maximum} assumes that the optimal (demonstrated) behavior is induced by maximizing an unknown value function. The goal then is to learn that value function, which imposes a certain structure onto the policy class.  The latter setting \cite{daume2009search,ross2011reduction,ho2016generative} makes no such structural assumptions and aims to directly fit a policy whose decisions well imitate the demonstrations.  This latter setting is typically more general but often suffers from higher sample complexity.  %In reinforcement learning, methods such as Q-learning fall into the former setting, and policy gradient methods fall into the latter \cite{sutton1998introduction}.
Our approach is agnostic to this dichotomy and can accommodate both styles of learning. Some instantiations of our framework allow for deriving theoretical guarantees, which rely on the policy learning setting.  %which does require committing to the policy learning paradigm.
Sample complexity comparison between imitation learning and reinforcement learning has not been studied much in the literature, perhaps with the exception of the recent analysis of AggreVaTeD~\cite{sun2017deeply}.

\textbf{Hierarchical Reinforcement Learning.} Feudal RL is another hierarchical framework that is similar to how we decompose the task hierarchically \citep{dayan1993feudal,dietterich2000hierarchical,vezhnevets2017feudal}.
In particular, a feudal system has a manager (similar to our \hi-level learner) and multiple submanagers (similar to our \lo-level learners), and submanagers are given pseudo-rewards which define the subgoals. Prior work in feudal RL use reinforcement learning for both levels; this can require a large amount of data when one of the levels has a long planning horizon, which we demonstrate in our experiments. In contrast, we propose a more general framework where imitation learners can be used to substitute reinforcement learners to substantially speed up learning, whenever the right level of expert feedback is available.   Hierarchical policy classes have been additional studied by \citet{he2010puma}, \citet{hausknecht2016deep}, \citet{zheng2016long}, and \citet{andreas2016modular}.

\textbf{Learning with Weaker Feedback.} Our work is motivated by efficient learning under weak expert feedback. When we only receive demonstration data at the high level, and must utilize reinforcement learning at the low level, then our setting can be viewed as an instance of learning under weak demonstration feedback.
The primary other way to elicit weaker demonstration feedback is with preference-based or gradient-based learning, studied by \citet{furnkranz2012preference}, \citet{loftin2016learning}, and \citet{christiano2017deep}.
\end{document}